\documentclass[letterpaper, 10 pt, journal]{IEEEtran}
\usepackage[utf8]{inputenc}
\usepackage{amsmath,bm} 
\usepackage{amssymb}
\usepackage{graphicx}
\usepackage{blindtext}
\usepackage{wrapfig}
\usepackage{color,soul}
\usepackage{cite}
\usepackage{svg}

\begin{document}

\title{Perfectly Undetectable Reflection and Scaling False Data Injection Attacks via Affine Transformation on Mobile Robot Trajectory Tracking Control}

\author{
Jun Ueda$^{1}$ and Hyukbin Kwon$^{1}$%
\thanks{$^{1}$Jun Ueda and Hyukbin Kwon are with the George W. Woodruff School of Mechanical Engineering, Georgia Institute of Technology, Atlanta, GA 30332-0405, USA. (e-mail: {\tt\footnotesize jun.ueda@me.gatech.edu, bin.kwon@gatech.edu})}
\thanks{This work was supported in part by the National Science Foundation under Grant No. 2112793.}%
}

\markboth{}
{Ueda \& Kwon : Perfectly undetectable false data injection attacks on mobile robot control} 

\maketitle

\begin{abstract}

With the increasing integration of cyber-physical systems (CPS) into critical applications, ensuring their resilience against cyberattacks is paramount.
A particularly concerning threat is the vulnerability of CPS to deceptive attacks that degrade system performance while remaining undetected.
This paper investigates perfectly undetectable false data injection attacks (FDIAs) targeting the trajectory tracking control of a non-holonomic mobile robot.
The proposed attack method utilizes affine transformations of intercepted signals, exploiting weaknesses inherent in the partially linear dynamic properties and symmetry of the nonlinear plant.
The feasibility and potential impact of these attacks are validated through experiments using a Turtlebot 3 platform, highlighting the urgent need for sophisticated detection mechanisms and resilient control strategies to safeguard CPS against such threats.
Furthermore, a novel approach for detection of these attacks called the state monitoring signature function (SMSF) is introduced.
An example SMSF, a carefully designed function resilient to FDIA, is shown to be able to detect the presence of a FDIA through signatures based on systems states.

\end{abstract}

\begin{IEEEkeywords}    
Mobile robots, False data injection attack, Affine transformation, Non-holonomic constraints, Nonlinear kinematics, Trajectory tracking, Stability, Security
\end{IEEEkeywords}

\section{Introduction}

 Virtually all current robotic systems are interconnected through computer networks for exchanging sensor measurements, control commands, and other information for monitoring and controlling purposes \cite{cyber-physical}. Mobile robots are examples of such systems and have become integral to a broad spectrum of applications, particularly in scenarios where human intervention is either impractical or inefficient. These applications range from industrial automation and logistics, where mobile robots handle materials, to exploration and data collection in hazardous environments such as deep-sea locations, disaster sites, and space missions \cite{Rubio19}.
 Given the increasing reliance on mobile robots for critical tasks and their operation in potentially unsecured or remote environments, ensuring the robustness of these systems against cyberattacks is an important area of research \cite{Jiang23}.

\begin{figure}[t]
	\centering
	\includegraphics[width=0.8\columnwidth]{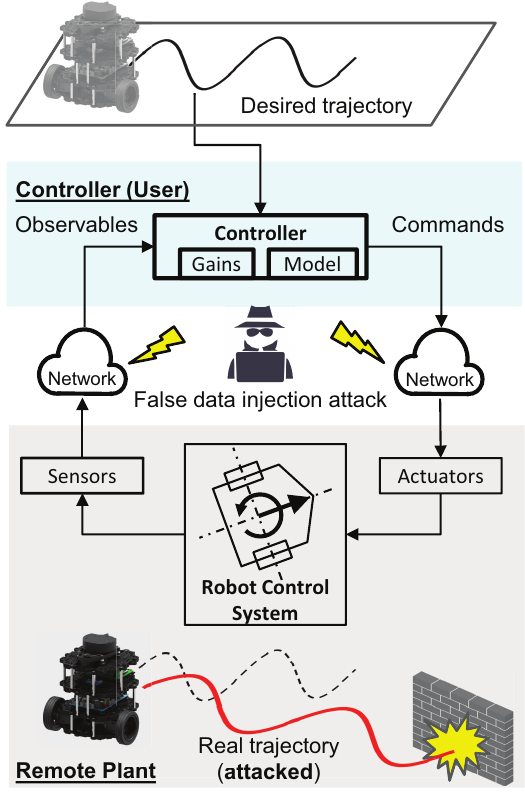}
	\caption{Conceptual diagram of false data injection attack (FDIA) on remote mobile robot control system.}
	\label{FDIA_mobilerobot_concept}
\end{figure}

The operation of these mobile robots often relies on networked communication systems to receive commands and transmit data back to the operators or control servers. This networked nature, while enabling remote and autonomous operations, is also susceptible to cybersecurity threats. One significant threat is False Data Injection Attacks (FDIAs)\cite{Li18,Jiang23,Dong20}, where an attacker manipulates the data being sent to or from the robot, or both, leading to incorrect actions, decision-making based on false information, or even taking control of the robot's operations.
For instance, in an FDIA, the data regarding the robot's location or sensor measurements could be compromised, misleading the navigation system and causing the robot to deviate from its intended path. In more sophisticated scenarios, as shown in Fig. \ref{FDIA_mobilerobot_concept}, attackers could inject false data to make the robot's system believe it is operating normally, referred to as undetectable or stealthy FDIAs \cite{Sandberg22,Mao20}, while it performs unintended tasks or causes physical damage to its surroundings. 

Stealthy and undetectable attacks are characterized by their increased difficulty for operators to detect. 
In stealthy attacks, an attacker capable of intercepting the original messages can inject the attack with partial or no knowledge of the plant, ensuring that the changes remain below the threshold of an attack detector \cite{Sandberg22}.
In the case of an undetectable attack, the attacked signals coincide with those that are within the regular operating range, causing faults and standard detectors to fail \cite{Pasqualetti13}. Perfectly undetectable attacks are those where there is no change in observed states, yet data integrity has been compromised. 
Similar attacks to those introduced in this paper have been discussed as covert attacks, as proposed in \cite{SMITH201190}, whereby if the attacker has perfect knowledge of the plant, it is possible to mask the attack from the controller's perspective.

Most works on covert attacks address linear time invariant \cite{SMITH201190,Li18,Pasqualetti13,ueda2024affine} systems.
FDIA attacks have been implemented to systems with moderate nonlinearities \cite{simplifiedFDIA}.
In this case the a simplified linearized version of the actual dynamics is used for the basis of the attack.
Other considerations of non-stealthy FDIA to nonlinear systems have been made to a class of nonlinear systems \cite{Wu23}.
In contrast, this paper specifically discusses perfectly undetectable FDIA applied to nonlinear mobile robot dynamics.

The robustness of closed-loop systems to account for uncertainties, disturbances, and sensor noise is a well-established and extensively studied field of research. 
Common compensation strategies from the control-theoretical standpoint include robust optimal control \cite{amin2009safe}, adaptive control\cite{petrillo2020secure }, state and disturbance compensation\cite{chen2015disturbance}. 
For anomaly detection associated with FDIA, a specific strategy involves a model-based control approach: the controller compares the observed plant behaviors induced by its control commands with those simulated based on a nominal plant dynamic model \cite{Pasqualetti13,Mao20,Sandberg22}. Any discrepancies identified through this process could indicate potential false-data injection, disturbances, or plant uncertainties. 
Encrypting communication lines or control algorithms \cite{EncryptedControlForNetworkedSystems, 10196122} adds another layer of protection. However, malleability of homomorphic encryption schemes can be exploited to apply FDIA \cite{TERANISHI2019297, boneh2012targeted}.
Conversely, if the controller observes little (i.e., stealthy) or no changes (i.e., perfectly undetectable) in the plant dynamics between normal and attacked states, model-based anomaly detection would be ineffective \cite{SMITH201190,Milošević20,GRACY21,Schellenberger}. Perfectly undetectable attacks are those in which there are no changes in the observed states, even though the closed-loop system is under attack and performs unintended motions.

The significance of this paper lies in the formulation of a generalized FDIA that involves coordinated multiplicative and additive data injections on both control commands and observables, taking form of affine transformations. Unlike covert attacks that require extensive computation and manipulation of the closed-loop system with complete knowledge of the plant dynamics \cite{Schellenberger}, this relatively simplistic and static FDIA allows attackers to execute perfectly undetectable attacks on remotely controlled mobile robots. 
Employing a classical two-wheel mobile robot kinematic model as a case study, this paper demonstrates how the inherent structure of commonly used nonlinear robot dynamics, from commands to outputs, enables a range of perfectly undetectable FDIAs. This vulnerability persists regardless of the type of trajectory control (e.g., \cite{Kanayama}), resulting in undetected failures within the controller's attack detection algorithms.

As a countermeasure, the paper proposes a state monitoring signature function (SMSF) approach along with an associated implementation architecture to continuously monitor for indications of perfectly undetectable FDIAs.
A signature function can be constructed from polynomial functions that are resilient to scaling and reflection attacks. While not permanently secure, the signature function can be designed to be difficult for the attacker to adversarially estimate for spoofing attacks. The SMSF approach differs from hash functions \cite{Merkle1989} and auxiliary systems \cite{Schellenberger}. SMSF operates on continuous and dynamic system states rather than static data as employed in hash functions. Additionally, SMSF can be implemented as a fully software solution, avoiding the need for an additional dynamic component often required by auxiliary systems.

This paper is organized as follows:
Section \ref{mobilerobotkinematics} provides preliminaries on well-known mobile robot dynamics and representative trajectory control methods. Section \ref{pefectlyundertectablefida} discusses perfectly undetectable FDIA on nonlinear control systems and introduces affine transformation-based formulations. Section \ref{solutiontoperfectlyundetectablefida} offers two solutions to the perfectly undetectable FDIA problem. Section \ref{stabilityfida} analyzes the stability of the closed-loop system under perfectly undetectable FDIA. Section \ref{experiments} presents experimental results. Section \ref{statemonitoring} introduces a SMSF as a countermeasure to perfectly undetectable FDIAs. While promising, this method has its own limitations.
Section \ref{discussion} discusses key observations and limitations, and Section \ref{conclusion} provides concluding remarks.

\section{Preliminaries on Mobile robot dynamic and control}
\label{mobilerobotkinematics}

\begin{figure}[t]
	\centering
	\includegraphics[width=0.8\columnwidth]{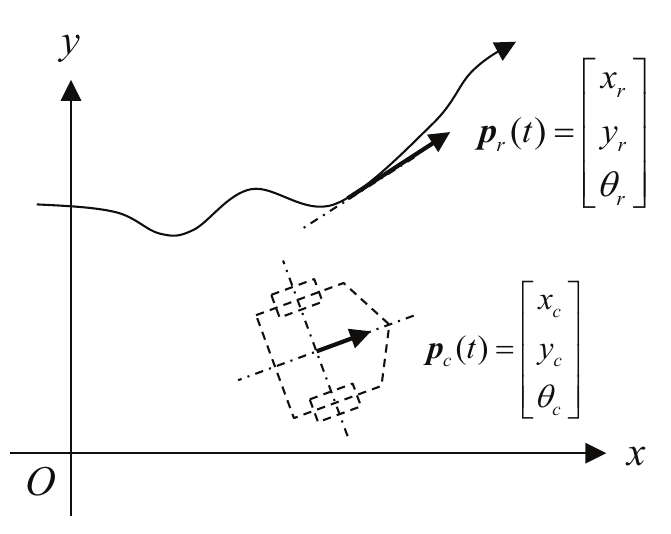}
	\caption{Mobile robot desired and current postures.
 The same notations are used as in \cite{Kanayama}.}
	\label{mobilerobotpostures}
\end{figure}

Well-known dynamic equations of a typical two-wheel mobile robot on a 2D plane (e.g., \cite{Kanayama}) are given below for readers' convenience. Interested readers can find extensive resources online and in the literature \cite{tzafestas2013introduction}.

The position of the robot can be represented with 3 degrees of freedom (DOF) as shown in Fig. \ref{mobilerobotpostures} where $x_c$ and $y_c$ are positions and $\theta_c$ is the orientation in the global frame: 

\begin{equation}
    \bm{p}_c = \begin{bmatrix}
    x_c\\
    y_c\\
    \theta_c\\        
    \end{bmatrix}.
    \label{eq:posture}
\end{equation}

\noindent 
The mobile robot can only be moved in 2 DOFs due to its non-holonomic constraints:
\begin{equation}
    \bm{q} = \begin{bmatrix}
    v\\
    \omega\\
    \end{bmatrix}
    \label{eq:vel}
\end{equation}
\noindent where $v$ and $\omega$ are the linear and angular velocities in the robot's local coordinate frame. 

A typical controller for tracking a given reference trajectory $\bm{r}(t) \in \mathcal{R}^{2\times 2}$ is implemented. 
The inputs to the controller are the reference posture $\bm{p}_r=[x_r, y_r, \theta_r]^T$ and the robot's current posture $\bm{p}_c$. Typically, both the error between the reference posture and current posture $\bm{p}_e=\bm{p}_r-\bm{p}_c$ as well as the reference linear and angular velocities $\bm{q}_r=[v_r, \omega_r]^T$ computed from $\bm{r}(t)$ are used. 

The dynamics of the mobile robot are given in (\ref{eq:kinematics}) as a first-order nonlinear equation:

\begin{equation}
    \dot{\bm{p}}_c = \bm{J}(\bm{p}_c) \bm{q},
        \label{eq:kinematics}
\end{equation}   
\noindent where $\bm{J}$ is the Jacobian matrix that maps the control command 
$\bm{q}$ onto the time derivative of $\bm{p}_c$:

\begin{equation}
    \bm{J}= \begin{bmatrix}
    \cos \theta_c  & 0 \\
    \sin \theta_c  & 0 \\
    0 & 1 \\
    \end{bmatrix}.
    \label{eq:jacobian}
\end{equation}

\noindent The controller outputs the input vector $\bm{q}$ as a control command that is sent via the communication channel.

One of the well-cited tracking control schemes was proposed by Kanayama \cite{Kanayama}, which this paper adopts as a representative control scheme:  
\begin{equation}
\bm{q} = \left[ {\begin{array}{*{20}{c}}
v\\
\omega 
\end{array}} \right] = \left[ {\begin{array}{*{20}{c}}
{{v_r}\cos {\theta _e} + {k_x}{x_e}}\\
{{\omega _r} + {v_r}({k_y}{y_e} + {k_\theta }\sin {\theta _e})}
\end{array}} \right],\label{kayanamacontrollaw}
\end{equation}
where the error ${\bm{q}_e} = 0$ was proven to be globally asymptotically stable with a Lyapunov function defined as:
$V = \frac{1}{2}({x_e}^2 + {y_e}^2) + (1 - \cos {\theta _e})/{k_y}$.
It should be noted that the attacker is not required to know the tracking control type or its gains to successfully implement a perfectly undetectable FDIA presented in this paper.

\section{Perfectly Undetectable FDIA: Coordinating Attacks on Observables and Control Commands}
\label{pefectlyundertectablefida}

\subsection{Fundamental equations}

Consider a general nonlinear dynamic plant in affine form:

\begin{eqnarray}
        {\bm{\dot x}} = \bm{f}({\bm{x}}) + \bm{g}({\bm{x}}){\bm{u}}
    \label{nonlineareqgeneralxdot}\\
{\bm{y}} = \bm{h}({\bm{x}}),
\label{nonlineareqgeneraly}
\end{eqnarray}
\noindent where $\bm{f}$ and $\bm{g}$ are both Lipschitz continuous functions, and a state feedback law is given by:

\begin{equation}
    \bm{u}=\bm{k}(\bm{x}) \label{nonlineareqgeneralu}.
\end{equation}

\begin{figure*}[t]
	\centering
	\includegraphics[width=1.7\columnwidth]{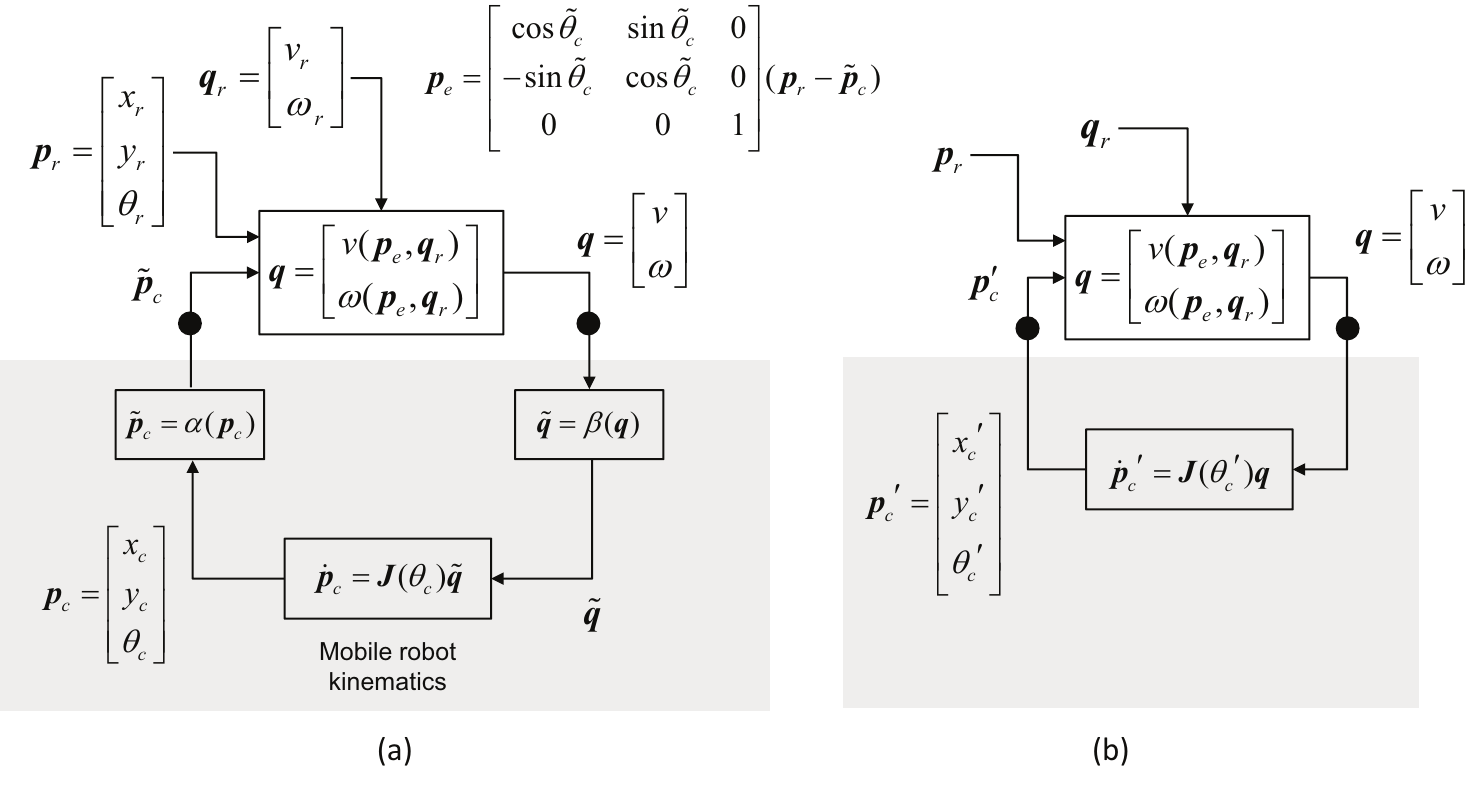}
	\caption{Perfectly undetectable false data injection attack (FDIA) on remote mobile robot control system based on affine transformations: (a) Attacked control system with coordinated FDIA on the commands and observables. (b) Plant dynamics as perceived by the controller, indistinguishable from the nominal robot behavior and thus undetectable.}
	\label{FDIA_mobilerobot}
\end{figure*}

It is assumed that a remote dynamic plant, described by  (\ref{nonlineareqgeneralxdot})(\ref{nonlineareqgeneraly}), is controlled via a network by a controller defined (\ref{nonlineareqgeneralu}). 
A generalized form of FDIA that involves coordinated multiplicative and additive data injections into both control commands and observables is depicted in Fig. \ref{FDIA_mobilerobot} (a), 

\begin{equation}
    {\bm{\tilde x}} = \bm{\alpha}(\bm{x}) ,
\end{equation}
\noindent where $\bm{\alpha}$ is a static observables attack function, and its inverse function $\bm{\alpha}^{-1}$ exits. Similarly, to the control command, $\tilde{\bm{u}}$ is a compromised (attacked) control command vector resulting from the attack by a static attack function, $\bm{\beta}$,

\begin{equation}
{\bm{\tilde u}} = \bm{\beta}(\bm{u}).
\end{equation}

Under the attack defined by $\bm{\alpha}$ and $\bm{\beta}$, the controller perceives the plant dynamics based on the command and the compromised observables, i.e., 
\begin{eqnarray}
    \dot{\tilde{\bm{x}}} 
    &=& \frac{\partial \bm{\alpha} (\bm{x})}{\partial \bm{x}} \dot{\bm{x}} \nonumber\\
    &=& \frac{\partial \bm{\alpha} (\bm{x})}{\partial \bm{x}} \bm{f}({\bm{x}}) + \bm{g}({\bm{x}}){\bm{\beta}(\bm{u})}\nonumber\\
    &=& \left. \frac{\partial \bm{\alpha} (\bm{x})}{\partial \bm{x}}\right|_{\bm{x}=\bm{\alpha}^{-1}(\bm{\tilde{x}})} \bm{f}(\bm{\alpha}^{-1}(\bm{\tilde{x}}))  + \bm{g}((\bm{\alpha}^{-1}(\bm{\tilde{x}})){\bm{\beta}(\bm{u})} . \nonumber \\
    \label{dottildex}
\end{eqnarray}

Following the nominal plant dynamics (\ref{nonlineareqgeneralxdot}) and (\ref{nonlineareqgeneraly}), let's define $\bm{x}'$ and $\bm{y}'$ that evolve with the same input $\bm{u}$ introduced by the attacker to mislead the controller into believing that the plant is operated normally:

\begin{eqnarray}
        {\bm{\dot x}'} &=& \bm{f}({\bm{x}'}) + \bm{g}({\bm{x}'}){\bm{u}}
    \label{nonlineareqgeneralxprimedot}\\
{\bm{y}'} &=& \bm{h}({\bm{x}'})
\label{nonlineareqgeneralyprime}\\
\bm{x}'(0)&=&\bm{x}(0).
\end{eqnarray}

{\bf Proposition 1: Indistinguishable plant responses amidst perfectly undetectable FDIA}: 
If $\bm{\alpha}$ and $\bm{\beta}$ exist such that the following conditions hold, a perfectly undetectable FDIA is achieved where $\bm{\tilde{x}}(t)=\bm{x}'(t), \forall t \geq 0$, {\it regardless} of the controller $\bm{k}(\bm{x})$.
\begin{itemize}
  \item Condition 1 (observing the nominal initial conditions): $\bm{x}(0)=\bm{\alpha}(\bm{x}(0))$  ensures that the observed state of the mobile robot at the start of the attack matches its actual state. This condition is crucial because if the attacker modifies the initial observed state, the controller would immediately detect a discrepancy and recognize the presence of an attack. In essence, the attack must begin by presenting the controller with the true initial state of the robot.

  \item Condition 2 (observing the nominal dynamics): 
    \begin{multline}
             \left. \frac{\partial \bm{\alpha} (\bm{x})}{\partial \bm{x}}\right|_{\bm{x}=\bm{\alpha}^{-1}(\bm{\tilde{x}})} \bm{f}(\bm{\alpha}^{-1}(\bm{\tilde{x}}))  + \bm{g}((\bm{\alpha}^{-1}(\bm{\tilde{x}})){\bm{\beta}(\bm{u})}\\
     = \bm{f}({\bm{x}'}) + \bm{g}({\bm{x}'}){\bm{u}}, \forall \bm{u}\\
     \label{condition2}
    \end{multline}
  (\ref{condition2}) ensures that the compromised system's dynamics, as observed by the controller, is identical to the nominal (unattacked) system's dynamics for all possible inputs, $\bm{u}$.
\end{itemize}

{\it Proof:}
This proposition is a direct corollary of the Picard–Lindelöf Theorem \cite{hartman2002ordinary}, or the uniqueness of the solution to an initial value problem for an ordinary differential equation.
The first condition must be satisfied for the controller to observe the same initial state:
$\tilde{\bm{x}}(0)=\bm{\alpha}(\bm{x}(0))=\bm{x}'(0)=\bm{x}(0)$, otherwise an attack detector in the controller would immediately detect a data falsification. Once the first condition is satisfied, 
$\bm{\tilde{x}}$, which evolves according to 
(\ref{dottildex}), and $\bm{\tilde{x}}'$, which evolves according to (\ref{nonlineareqgeneralxprimedot}) yield identical values at all times when the same $\bm{u}$ is applied. $\blacksquare$

\begin{figure}[t]
	\centering
	\includegraphics[width=0.9\columnwidth]{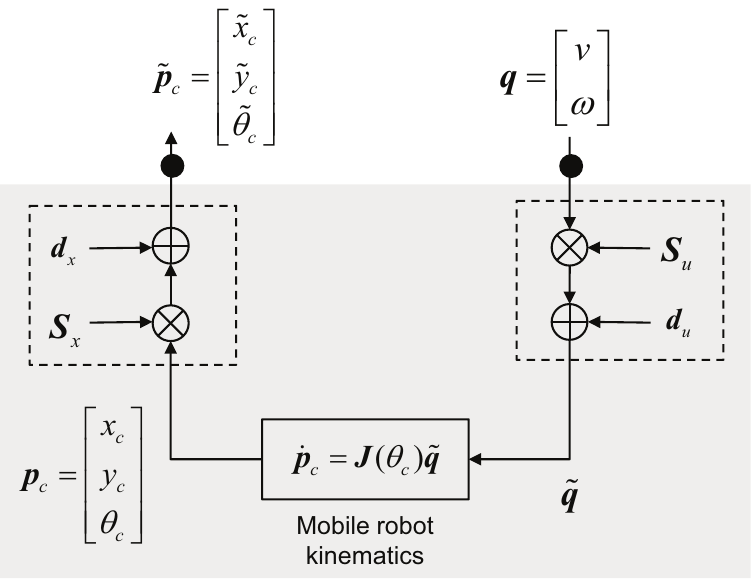}
	\caption{Affine transformation-based coordinated FDIA on mobile robot control system}
	\label{mobilerobot_affinetransformation_FDIA}
\end{figure}

\subsection{Specific conditions of perfectly undetectable FDIA on  mobile robot control}
The mobile robot dynamic equation (\ref{eq:kinematics}) is a special case of  (\ref{nonlineareqgeneralxdot}) and (\ref{nonlineareqgeneraly}) where

\begin{eqnarray}
\bm{x} &=& {{\bm{p}}_c}\\
\bm{u} &=& {\bm{\tilde q}}\\
f({\bm{x}}) &=& \bm{0} \label{fx=0}\\
g({\bm{x}}) &=& {\bm{J}}({\theta _c}) = \left[ {\begin{array}{*{20}{c}}
{\cos {\theta _c}}&0\\
{\sin {\theta _c}}&0\\
0&1
\end{array}} \right]\\
h({\bm{x}}) &=& {\bm{x}} = {{\bm{p}}_c}.
\end{eqnarray}

\noindent The attack functions are also defined as:

\begin{equation}
    {\bm{\tilde p}_c} = \bm{\alpha}(\bm{p}_c) ,
\end{equation}
\begin{equation}
{\bm{\tilde q}} = \bm{\beta}(\bm{q}).
\label{tildeu}
\end{equation}
\noindent 
The dynamic relationship between the command $\bm{q}$ and the observable $\tilde{\bm{p}}_c$ illustrated as a shaded region in 
Fig. \ref{FDIA_mobilerobot} (a) is given as:

\begin{equation}
    \dot{\tilde{\bm{p}}}_c 
    = \frac{\partial \bm{\alpha} (\bm{p}_c)}{\partial \bm{p}_c} \dot{\bm{p}}_c 
    = \frac{\partial \bm{\alpha} (\bm{p}_c)}{\partial \bm{p}_c} \bm{J}(\theta _c){\tilde{\bm{q}}} 
    = \frac{\partial \bm{\alpha} (\bm{p}_c)}{\partial \bm{p}_c} \bm{J} (\theta _c) \bm{\beta} (\bm{q}).
\end{equation}

When the controller perceives the attacked plant dynamics as matching the nominal plant dynamics, a perfectly undetectable FDIA is considered to be achieved, as illustrated in Fig. \ref{FDIA_mobilerobot} (b), i.e.,

\begin{equation}
\dot{\tilde{\bm{p}}}_c  = \frac{\partial h (\bm{p}')}{\partial \bm{p}'} \dot{\bm{p}}' =\bm{J}(\theta' _c)\bm{q},
\label{dottildepc}
\end{equation}
\noindent where $\bm{p}'=[x'_c, y'_c, \theta'_c]$ is a {\it fake} state variable vector. In fact, $\bm{p}' \neq \bm{p}$, and the excepted behavior is different from the actual behavior.
If the controller observes
\begin{equation}
\bm{p}'=\tilde{\bm{p}}_c = \bm{\alpha} (\bm{p}_c) 
\label{pcprime}
\end{equation}
\noindent from (\ref{dottildepc}) and (\ref{pcprime}), the observed dynamics by the controller becomes equivalent to
\begin{equation}
\dot{\bm{p}}'= \bm{J}(\theta'_c)\bm{q}
\label{dotpprime}
\end{equation}
which matches with the nominal dynamics (\ref{eq:kinematics}), achieving an perfectly undetectable FDIA, {\it regardless} of the controller that generates $\bm{q}$. In a later section, a theorem with conditions including one about the initial conditions will be given.

\subsection{Problem formulation using affine transformation-based FDIA}

The main objective of the paper is to discuss the existence of attack functions $\bm{\alpha}$ and $\bm{\beta}$ for (\ref{eq:kinematics}) that yield 
(\ref{dotpprime}). 
For simplicity, let's assume that the attacker opts for a linear affine transformation as shown in Fig. \ref{mobilerobot_affinetransformation_FDIA} instead of general nonlinear attack functions. 
This assumption is not entirely unrealistic. When the communication lines are encrypted using homomorphic encryption algorithms (HE), they impose limited computational capabilities on the attacker\cite{ueda2024affine}, allowing only simple operations such as multiplication and addition to be performed on the original messages in the communication lines. This type of vulnerability is known as a malleability attack \cite{TERANISHI2019297} \cite{boneh2012targeted}.

The affine transformation attack to the observables is given as follows:

\begin{equation}
    {\bm{\tilde p}_c} = \bm{\alpha}(\bm{p}_c) =\bm{S}_x \bm{p}_c + {{\bm{d}}_x}, \label{tildepc}
\end{equation}
\noindent or, alternatively, in the form of homogeneous transformation:
\begin{equation}
\left[ {\begin{array}{*{20}{c}}
{{\bm{\tilde p}_c}}\\
1
\end{array}} \right] = \left[ {\begin{array}{*{20}{c}}
{{{\bm{S}}_x}}&{{{\bm{d}}_x}}\\
0&1
\end{array}} \right]\left[ {\begin{array}{*{20}{c}}
{\bm{p}_c}\\
1
\end{array}} \right]
\end{equation}
\noindent
where $\bm{S}_x \in \mathbb{R}^{n \times n}$ represents an arbitrary transformation, such as scaling, shear, and rotation, and  $\bm{d}_x \in \mathbb{R}^n$ represents a translation introducing an offset. This paper assumes that $\bm{S}_x$ and $\bm{d}_x$ are constants. 

Similarly, as with the control command, $\bm{S}_u \in \mathbb{R}^{m \times m}$ represents an arbitrary transformation and $\bm{d}_u \in \mathbb{R}^m$ represents a translation introducing an offset. This paper assumes that $\bm{S}_u$ and $\bm{d}_u$ are constants:

\begin{equation}
{\bm{\tilde q}} = \bm{\beta}(\bm{q})={{\bm{S}}_u}{\bm{q}} + {{\bm{d}}_u},
\label{tildeq}
\end{equation}
\begin{equation}
\left[ {\begin{array}{*{20}{c}}
{{\bm{\tilde q}}}\\
1
\end{array}} \right] = \left[ {\begin{array}{*{20}{c}}
{{{\bm{S}}_u}}&{{{\bm{d}}_u}}\\
0&1
\end{array}} \right]\left[ {\begin{array}{*{20}{c}}
{\bm{q}}\\
1
\end{array}} \right]
\end{equation}

In the literature, most studies considered either additive or multiplicative FDIA on control commands or observables. For example, Zhu 2023 \cite{Zhu23} studied both multiplicative and additive data injections, assuming that the observables remained uncompromised. 
Representing FDIAs in the form of affine transformations with (\ref{tildepc}) and (\ref{tildeq}) allow for more generalized analyses. 
It should be mentioned that simultaneous FDIA on the commands and observables is not necessarily a new concept. Past works on covert attacks introduced a similar structure in which the attacker implements an additional dynamic controller between the commands and observables \cite{Schellenberger}. In contrast, this paper formulates perfectly undetectable FDIAs in terms of affine transformations, representing, to the authors' knowledge, for the first time this has been done on nonlinear robot system dynamics. 

Based on the aforementioned analysis, and {\bf Proposition 1}, the perfectly undetectable FDIA problem, which is specific to the mobile robot dynamics, can be defined as follows. 

{\bf Definition 1: Perfectly undetectable FDIA problem on mobile robot dynamics}. For the nominal plant dynamic equation (\ref{eq:kinematics}) with the Jacobian matrix (\ref{eq:jacobian}), if $\bm{S}_x$, $\bm{d}_x$, $\bm{S}_u$, and $\bm{d}_u$ exist such that 'fake' state variables $\bm{p}'$ can be defined and the following conditions hold, a perfectly undetectable FDIA is implemented. 

\begin{itemize}
    \item Condition 1 (Same initial condition): ${{\tilde{\bm{p}}}_c}(0) = {{\bm{S}}_x}{{\bm{p}}_c}(0) + {{\bm{d}}_x}={{\bm{p}}_c}(0)$
    \item Condition 2 (Same observed dynamics): The observed plant dynamics by the controller
$\dot{\bm{\tilde{p}}}_c 
= \bm{S}_x \bm{J} (\theta _c) (\bm{S}_u \bm{q} + \bm{d}_u)= \bm{S}_x \dot{\bm{p}_c}$ is equivalent to the 
nominal dynamics
$\dot{\bm{p}}'= \bm{J}(\theta'_c)\bm{q}$ evolved by the same command $\bm{q}(t)$
where the attacked dynamics is given by
$\dot{\bm{p}_c}= \bm{J}(\theta_c)(\bm{S}_u \bm{q} + \bm{d}_u)$ and $\theta_c = [0 \; 0 \; 1] \bm{p}_c$. In short, $\bm{J}(\theta_c)(\bm{S}_u \bm{q} + \bm{d}_u) = \bm{J}(\theta'_c)\bm{q}$ must be satisfied. 
\end{itemize}
\noindent
The next section will provide specific solutions to this problem.

\section{Solution to the perfectly undetectable FDIA problem}
\label{solutiontoperfectlyundetectablefida}

\begin{figure*}[th]
	\centering
	\includegraphics[width=1.7\columnwidth]{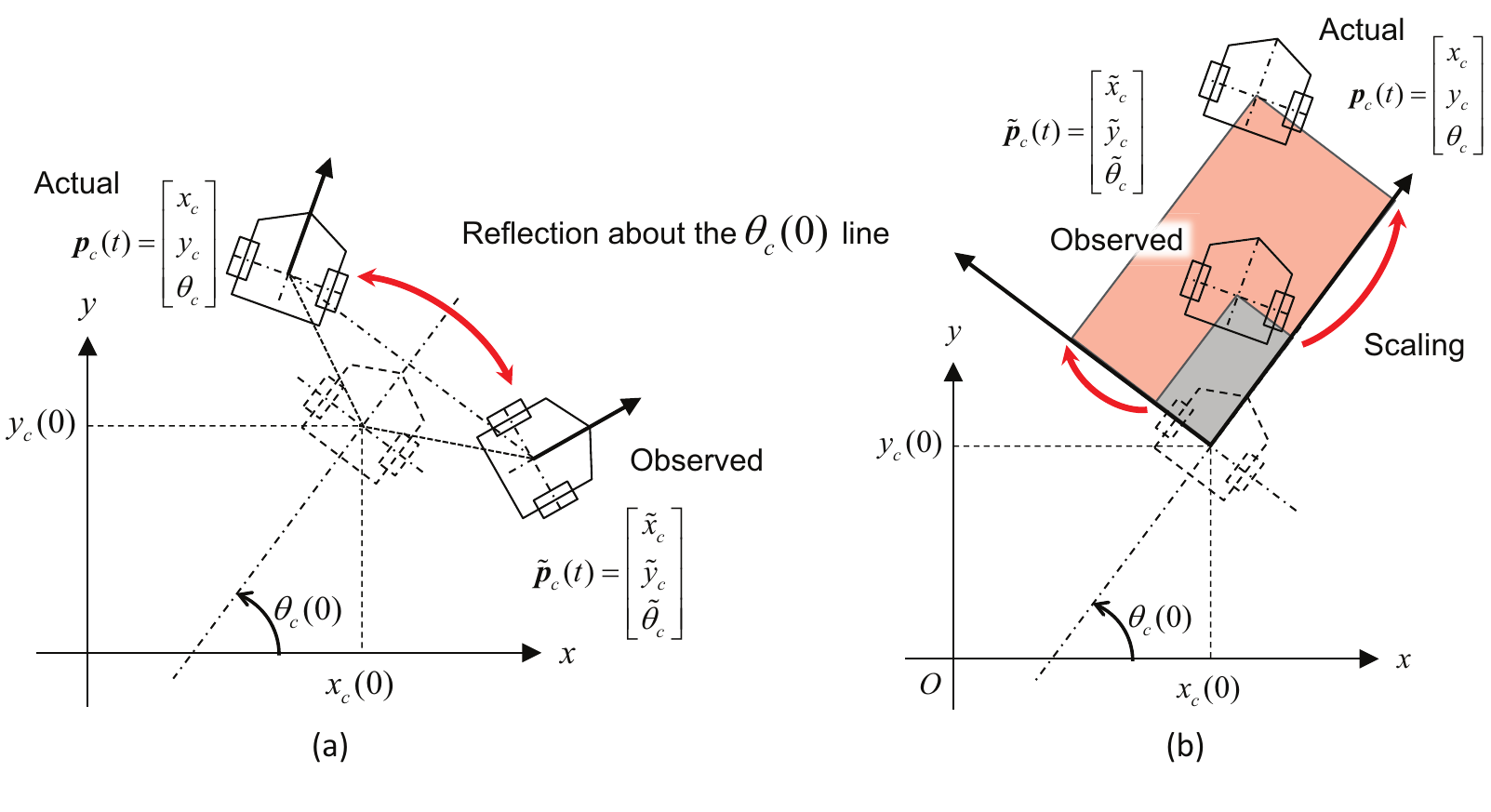}
	\caption{Perfectly undetectable FDIA solutions: (a) reflection and (b) scaling attacks}
	\label{mobilerobot_reflectionscaling}
\end{figure*}

\subsection{Attackability analysis of mobile robot Jacobian matrix}

Equation (\ref{eq:jacobian}) reveals a block-diagonal structure, with its (3,1) element being constant and decoupled from the (1,1) and (2,1) elements. This indicates that the robot's angle $\theta_c$ is governed by a first-order linear equation that solely depends on the input $\omega$. In this section, we first consider the structure of $\bm{S}_u$. Assuming $\bm{S}_u = \left[ {\begin{array}{*{20}{c}}
\beta_{11}& \beta_{12}\\
\beta_{21}&\beta_{22}
\end{array}} \right]$, the following Proposition is obtained. 

{\bf Proposition 2: General form of $\bm{S}_u$ and associated requirements}. The general form of $\bm{S}_u$ is given as a diagonal form, 
$\bm{S}_u = \left[ {\begin{array}{*{20}{c}}
\beta_{11}&0\\
0&{\pm 1}
\end{array}} \right]$ subject to the following requirements:
\begin{enumerate}
    \item  $\beta_{12}=0$ and $\beta_{21}=0$ under the assumption that $\bm{d}_x$ is a constant.
    \item  $\beta_{22}= \pm 1$.  
\end{enumerate}

{\it Proof:}
Consider the evolution of $\tilde{\bm{p}}_c =\int_0^t \bm{S}_u \bm{q} \;  \mathrm{d}t$, i.e., 
\begin{eqnarray}
\tilde{x}_c &=& \beta_{11} \int_0^t v \cos(\theta_c) \;  \mathrm{d}t + \beta_{12} \int_0^t \omega \cos(\theta_c) \;  \mathrm{d}t \nonumber \\
&=& \beta_{11} x_c -\beta_{12} \sin(\theta_c), \label{tildexwithbeta12}\\
\tilde{y}_c &=&    \beta_{11} y_c + \beta_{12} \cos(\theta_c),  \label{tildeywithbeta12}\\
\tilde{\theta}_c &=& \beta_{21}  \int_0^t v \; \mathrm{d}t +  \beta_{22} {\theta}_c .\label{tildethetawithbeta21}
\end{eqnarray}
\noindent
The second term in both (\ref{tildexwithbeta12}) and (\ref{tildeywithbeta12}) is a nonlinear function of $\theta_c$ and thus time-dependent. Unless $\bm{d}_x$ is computed from $\theta_c$ in real-time, the attacker cannot eliminate this term to realize a perfectly undetectable FDIA. 
Similarly, the second term in (\ref{tildethetawithbeta21}) is the length of the path produced by the robot. $\bm{p}_c$ does not store such information. Unless $\bm{d}_x$ includes an integration of $v$ over time, the attacker cannot eliminate this term to realize a perfectly undetectable FDIA. These observations contradict the assumption of a linear affine transformation, leading to $\beta_{12}=\beta_{21}=0$. 
Regarding (\ref{tildethetawithbeta21}), since $\beta_{21}=0$, $\tilde{\theta}_c =  \beta_{22} {\theta}_c$. See
Appendix \ref{appen_attacksincos} for Proposition B1 about the vulnerability of trigonometric functions. Considering $\cos(\tilde{\theta}_c)=\cos(\beta_{22} \theta_c)$, only $\beta_{22} = \pm 1$ is feasible. $\blacksquare$

{\bf Remark 1: Possible FDIA scenarios}. 
$\beta_{11}$ is a scaling factor that represents an attack on the linear velocity, termed a scaling attack. 
No attack is imposed on the linear velocity when $\beta_{11}=1$. Also, $\beta_{11}=0$ cannot be chosen since such an attack would be immediately detected by the controller, thus $\beta_{11}\neq 0$.  
Since no attack is imposed on the angular velocity command when $\beta_{22}=1$ (i.e., the trivial case), the only effective selection of $\beta_{22}=-1$, a scenario, termed a reflection attack.

{\bf Remark 2: Future time-variant $\bm{d}_x$}. $\beta_{12} \neq 0$ and $\beta_{21} \neq 0$ may be used when $\bm{d}_x$ is time-variant. This consideration is beyond the scope of this particular paper and will be addressed in future work.

\subsection{Main result: perfectly undetectable FDIA solutions}
\label{FDIAsolutions}

Based on the aforementioned analysis, the following theorem is obtained that shows the existence of specific solutions for affine transformation based perfectly undetectable FDIAs.

{\bf Theorem 1: Specific FDIA solutions to mobile robot dynamics}
(See {\bf Proposition 1}).
\begin{itemize} 
    \item {\bf Condition 1}: $({{\bm{I}}_2} - {{\bm{S}}_x}){{\bm{p}}_c}(0) = {{\bm{d}}_x}$
    \item {\bf Condition 2}: $\bm{d}_u=0$ and
    \begin{itemize}
    \item  Condition 2-1: Reflection attack. \begin{equation}
    {{\bm{S}}_x} = \left[ \! \! {\begin{array}{*{20}{c}}
\frac{1}{\beta_{11}} {\cos (2{\theta _c}(0))}&\frac{1}{\beta_{11}} {\sin (2{\theta _c}(0))}&0\\
\frac{1}{\beta_{11}} {\sin (2{\theta _c}(0))}& - \frac{1}{\beta_{11}} {\cos (2{\theta _c}(0))}&0\\
0&0&{ - 1}\end{array}} \! \! \right]\\
\end{equation} and $\bm{S}_u = \left[ {\begin{array}{*{20}{c}}
\beta_{11}&0\\
0&{ - 1}
\end{array}} \right]$, where $\beta_{11}\neq 0$ (see {\bf Remark 1}), termed a {\it reflection} attack, 
\item Condition 2-2: Scaling attack.
\begin{equation}
    {{\bm{S}}_x} = \left[ \! \! {\begin{array}{*{20}{c}}
\frac{1}{\beta_{11}} & 0  &0\\
0 &  \frac{1}{\beta_{11}}  &0\\
0&0&{  1}\end{array}} \! \! \right]
\end{equation} and $\bm{S}_u = \left[ {\begin{array}{*{20}{c}}
\beta_{11}&0\\
0&{  1}
\end{array}} \right]$, termed a {\it scaling} attack.
\end{itemize}
\end{itemize}

{\it Proof:}
Note that the observation at $t=0$ must be unchanged, i.e., ${{\bm{p}}_c}(0) = {{\bm{S}}_x}{{\bm{p}}_c}(0) + {{\bm{d}}_x}$, Condition 1, 
\begin{equation}
    ({{\bm{I}}_2} - {{\bm{S}}_x}){{\bm{p}}_c}(0) = {{\bm{d}}_x},
\end{equation}
is obtained.

A possible attack may be a reflection attack on the angular velocity, i.e., 
$\bm{S}_u = \left[ {\begin{array}{*{20}{c}}
\beta_{11}&0\\
0&{ \pm 1}
\end{array}} \right]$,

\begin{equation}
   \bm{\beta} (\bm{q}) = \bm{S}_u \bm{q} + \bm{d}_u = \left[ {\begin{array}{*{20}{c}}
\beta_{11}&0\\
0&{ \pm 1}
\end{array}} \right]\left[ {\begin{array}{*{20}{c}}
v\\
\omega 
\end{array}} \right] + {{\bm{d}}_u} = \left[ {\begin{array}{*{20}{c}}
\beta_{11} v\\
{ \pm \omega }
\end{array}} \right] + {{\bm{d}}_u}, 
\end{equation}
yielding,
\begin{eqnarray}
\dot{\bm{\tilde{p}}}_c &=& \frac{\partial \bm{\alpha} (\bm{p}_c)}{\partial \bm{p}_c} \bm{J} (\theta _c) \bm{\beta} (\bm{q}) \nonumber \\
&=& \frac{\partial \bm{\alpha} (\bm{p}_c)}{\partial \bm{p}_c} \bm{J} (\theta _c) (\bm{S}_u \bm{q} + \bm{d}_u)
\nonumber \\
&=& \frac{{\partial \alpha ({{\bm{p}}_c})}}{{\partial {{\bm{p}}_c}}}\left[ {\begin{array}{*{20}{c}}
\beta_{11} {\cos {\theta _c}}&0\\
\beta_{11} {\sin {\theta _c}}&0\\
0&{ \pm 1}
\end{array}} \right]{\bm{q}} + \frac{{\partial \bm{\alpha} ({{\bm{p}}_c})}}{{\partial {{\bm{p}}_c}}}{\bm{J}}({\theta _c}){{\bm{d}}_u}. \nonumber \\
\end{eqnarray}

Note that the second term that works as a bias must be 
$\frac{{\partial \bm{\alpha} ({{\bm{p}}_c})}}{{\partial {{\bm{p}}_c}}}{\bm{J}}({\theta _c}){{\bm{d}}_u} = 0$
to observe the nominal dynamics by the controller. Since ${\bm{J}}({\theta _c})$ is state-dependent and time-variant, the attacker must choose $\bm{d}_u=\bm{0}$.

If
${\bm{J}}({\theta '_c}) = \frac{{\partial \bm{\alpha} ({{\bm{p}}_c})}}{{\partial {{\bm{p}}_c}}}\left[ {\begin{array}{*{20}{c}}
{\beta_{11} \cos {\theta _c}}&0 \nonumber\\
{\beta_{11} \sin {\theta _c}}&0\\
0&{ \pm 1}
\end{array}} \right]$,
the perfectly undetectable FDIA is successfully implemented.
Consider when $\beta_{22}=-1$ (reflection attack). 
Since
$\frac{{\partial \bm{\alpha} ({{\bm{p}}_c})}}{{\partial {{\bm{p}}_c}}} = {{\bm{S}}_x}$,

\begin{eqnarray}
{\bm{J}}({\theta '_c}) = \left[ {\begin{array}{*{20}{c}}
{ \cos \theta {'_c}}&0\\
{ \sin \theta {'_c}}&0\\
0&1
\end{array}} \right] = {{\bm{S}}_x}\left[ {\begin{array}{*{20}{c}}
{\beta_{11} \cos {\theta _c}}&0\\
{\beta_{11} \sin {\theta _c}}&0\\
0&{-1}
\end{array}} \right] \nonumber \\
= \left[ {\begin{array}{*{20}{c}}
{\left. {\underline {\, 
\frac{1}{\beta_{11}} {\bm{R}_{ref}({\theta _c} \to {{\theta '}_c})} \,}}\! \right| }&{\begin{array}{*{20}{c}}
0\\
0
\end{array}}\\
{\begin{array}{*{20}{c}}
0&0
\end{array}}&{ - 1}
\end{array}} \right]\left[ {\begin{array}{*{20}{c}}
{\beta_{11} \cos {\theta _c}}&0\\
{\beta_{11} \sin {\theta _c}}&0\\
0&{ - 1}
\end{array}} \right] \nonumber \\
= \left[ {\begin{array}{*{20}{c}}
{\left. {\underline {\, 
\frac{1}{\beta_{11}} {\bm{R}_{ref}({\theta _c} \to {{\theta '}_c})\left[ {\begin{array}{*{20}{c}}
{\beta_{11} \cos {\theta _c}}\\
{\beta_{11} \sin {\theta _c}}
\end{array}} \right]} \,}}\! \right| }&{\begin{array}{*{20}{c}}
0\\
0
\end{array}}\\
{\begin{array}{*{20}{c}}
0&0
\end{array}}&1
\end{array}} \right]
\end{eqnarray}
\noindent where
$\bm{R}_{ref}({\theta _c} \to {\theta'_c})$ is a reflection matrix to reflect $\theta_c$ to  $\theta'_c=\tilde{\theta}_c$ about the $\theta_c(0)$ line, as shown in Fig. \ref{mobilerobot_reflectionscaling}(a),  which is given as follows:
\begin{eqnarray}
\cos \theta {'_c} &\!\!=&\!\! \cos ( - {\theta _c} + 2{\theta _c}(0)) \nonumber \\
 &\!\!=&\!\! \cos ({\theta _c})\cos (2{\theta _c}(0)) + \sin ({\theta _c})\sin (2{\theta _c}(0))  \\
\sin \theta {'_c} &\!\!=&\!\! \sin ( - {\theta _c} + 2{\theta _c}(0)) \nonumber \\
 &\!\!=&\!\!  - \sin ({\theta _c})\cos (2{\theta _c}(0)) + \cos ({\theta _c})\sin (2{\theta _c}(0))
 \end{eqnarray}
 \begin{eqnarray}
\left[ {\begin{array}{*{20}{c}}
{\cos \theta {'_c}}\\
{\sin \theta {'_c}}
\end{array}} \right] = \bm{R}_{ref}({\theta _c} \to {{\theta '}_c})\left[ {\begin{array}{*{20}{c}}
{\cos {\theta _c}}\\
{\sin {\theta _c}}
\end{array}} \right]\\
\therefore \bm{R}_{ref}({\theta _c} \to {{\theta '}_c}) = \left[ {\begin{array}{*{20}{c}}
{\cos (2{\theta _c}(0))}&{\sin (2{\theta _c}(0))}\\
{\sin (2{\theta _c}(0))}&{ - \cos (2{\theta _c}(0))}
\end{array}} \right]
\end{eqnarray}
yielding, 
\begin{equation}
    {{\bm{S}}_x} = \left[ {\begin{array}{*{20}{c}}
\frac{1}{\beta_{11}} {\cos (2{\theta _c}(0))}&\frac{1}{\beta_{11}} {\sin (2{\theta _c}(0))}&0\\
\frac{1}{\beta_{11}} {\sin (2{\theta _c}(0))}& - \frac{1}{\beta_{11}} {  \cos (2{\theta _c}(0))}&0\\
0&0&{ - 1}\end{array}} \right].
\end{equation}
\noindent
Similarly, when $\beta_{22}=1$, $\bm{S}_u$ only imposes a scaling attack without reflection as illustrated in Fig. \ref{mobilerobot_reflectionscaling}(b), i.e., 
\begin{equation}
    {{\bm{S}}_x} = \left[ {\begin{array}{*{20}{c}}
\frac{1}{\beta_{11}} & 0  &0\\
0 &  \frac{1}{\beta_{11}}  &0\\
0&0&{  1}\end{array}} \right].
\end{equation}
$\blacksquare$

{\bf Remark 3: Initial conditions $\bm{p}_c(0)$ required by the attacker}. $\bm{p}_c (0)$ must be known by the attacker to satisfy Condition 1 at the onset of the attack. To relax this condition, time-variant attack parameters must be implemented that will be discussed in our future paper. 

{\bf Remark 4: Necessity of $\bm{d}_u=\bm{0}$}. 
Additive FDIA on control commands, $\bm{d}_u \neq \bm{0}$, is detectable and thus relatively easily compensated for by using traditional robust control methods such as disturbance observers. 
This is required primarily due to the inertial property of the robot dynamics without an explicit static equilibrium shown in (\ref{fx=0}). Conversely, for other dynamic systems with a non-zero drift vector field term, $f(\bm{x}) \neq \bm{0}$, a non-zero $\bm{d}_u$ may need to be determined.

\subsection{Stability of the closed-loop system with perfectly undetectable FDIAs}
\label{stabilityfida}
Recall that {\bf Proposition 1} indicates that the attacked system will remain convergent as long as a perfectly undetectable FDIA is implemented under a stabilizing controller such as (\ref{kayanamacontrollaw}), {\it regardless} of the specific controller used. Nevertheless, this section provides a sketch of proof to confirm the stability of the attacked system for a specific control scheme.

{\bf Proposition 3: Stability of trajectory tracking control (Kanayama \cite{Kanayama} modified)}. For the control scheme that uses the compromised observables ${\tilde x_e},{\tilde y_e},{\tilde \theta _e}$ due to FDIA, 
\begin{equation}
\bm{q} = \left[ {\begin{array}{*{20}{c}}
v\\
\omega 
\end{array}} \right] = \left[ {\begin{array}{*{20}{c}}
{{v_r}\cos {\tilde{\theta} _e} + {k_x}{\tilde{x}_e}}\\
{{\omega _r} + {v_r}({k_y}{\tilde{y}_e} + {k_\theta }\sin {\tilde{\theta} _e})}
\end{array}} \right],\label{kayanamacontrollawtilde}
\end{equation}
${\tilde{\bm{q}}_e} = 0$  is a stable equilibrium for the reference velocity $v_r>0$.

{\it Sketch of proof:} Since a perfectly undetectable FDIA is implemented, 
$\dot{\bm{\tilde p}}_c = {{\bm{S}}_x}{\bm{J}}({\theta _c}){{\bm{S}}_u}{\bm{q}} = {\bm{J}}({\theta}'){\bm{q}}$ holds. $\bm{\tilde p}_c$ evolves in exactly the same way as $\bm{p}'$ does. Therefore, we can perform the change of variables for the Jacobian matrix, i.e., $\dot{\bm{\tilde p}}_c = {\bm{J}}({\tilde{\theta} _c}){\bm{q}}$.
Consequently, the error dynamics associated with the control scheme can be fully expressed in terms of ${\tilde x_e},{\tilde y_e},{\tilde \theta _e}$. Likewise, for a
Lyapunov function candidate defined as:
$\tilde{V} = \frac{1}{2}({\tilde{x}_e}^2 + {\tilde{y}_e}^2) + (1 - \cos {\tilde{\theta} _e})/{k_y}$, its time derivative can be expressed by:
\begin{equation}
    \dot{\tilde{V}} = - {k_x}{\tilde x_e}^2 - {v_r}{k_\theta}{\sin ^2}\tilde \theta /{k_y} \le 0,
\end{equation}
\noindent
resulting in the same conclusion shown in \cite{Kanayama}. Since $\tilde{\bm{p}}_e(0) = {\bm{p}_e}'(0)$,  the error dynamics between the observed plant and that of the nominal plant match exactly, confirming a perfectly undetectable FDIA. $\blacksquare$

\section{Experiments}
\label{experiments}
\subsection{Mobile robot experimental setup}

\begin{figure}[t]
    \centering
    \includegraphics[width=1\columnwidth]{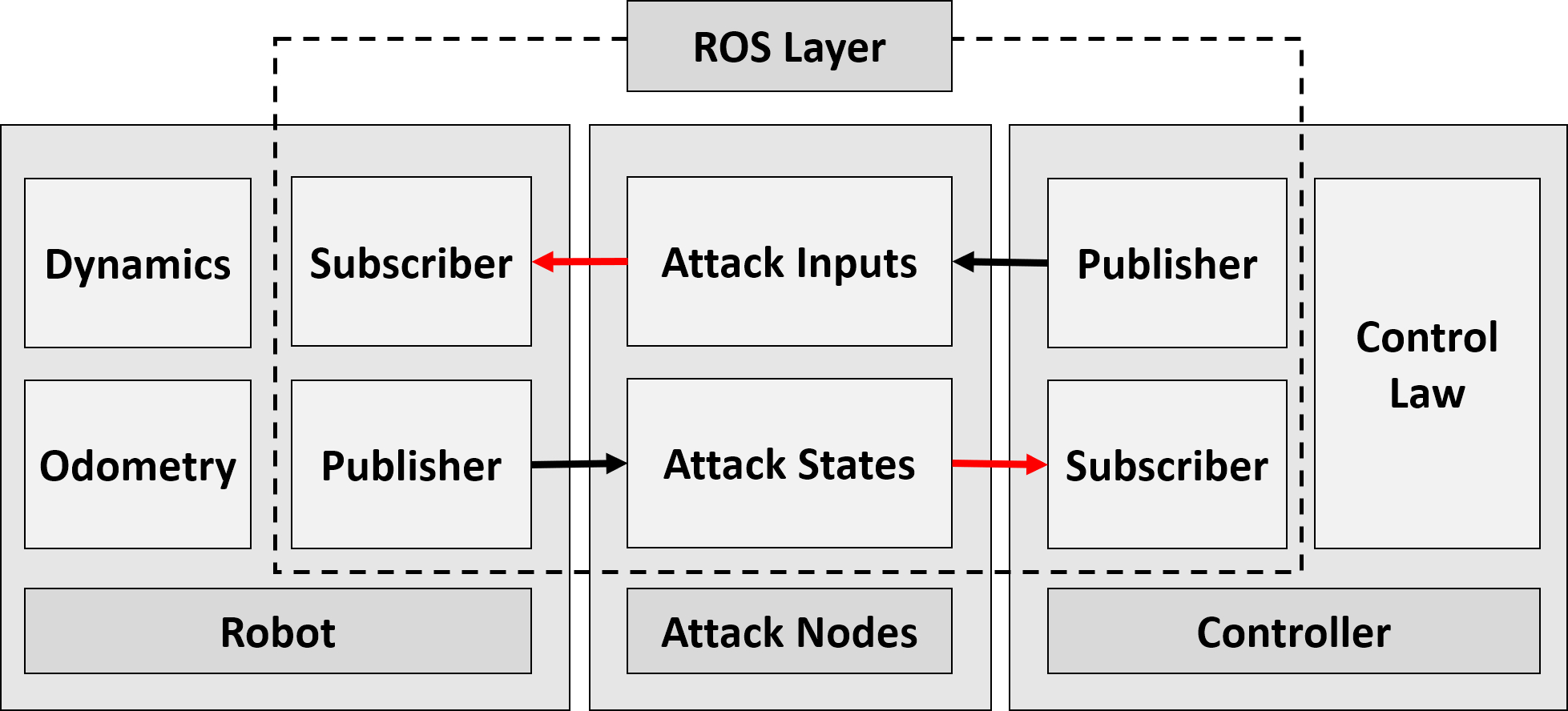} 
    \caption{Implemented ROS nodes for experimentation. Red arrows denote modified data.}
    \label{fig:ROSDesign}
\end{figure}

A non-holonomic mobile robot (Turtlebot 3) with an onboard computer (Raspberry Pi 3) and a separate computer running Ubuntu Linux (11th Gen Intel(R) Core(TM) i7-1165G7) functioning as the controller were used.
Communication between the robot and the remote controller was established using TCP/IP with ROS 2.
The design of the ROS network is shown in Fig. \ref{fig:ROSDesign}.
Each attacker node modifies the published inputs and observables according to the preloaded attack scenario.
Modified data shown in red arrows are received by the robot and the controller respectively.
Plant and controller nodes subscribe to the modified messages for use in the control loop.
The computer spins the controller and attacker nodes, while an onboard single-board computer on the robot listens to the input commands and broadcasts its current state.
The robot used Google Cartographer \cite{Wolfgang16} for localization during the task.
The controller node implements the controller presented in (\ref{kayanamacontrollawtilde}) \cite{Kanayama} with gains $K_x = 2$, $K_y = 2000$, and $K_\theta = 100$.
The controller was evaluated at 100 Hz, while errors and control inputs were logged at 50 Hz.

In order to satisfy Condition 1, the attacker was assumed to have knowledge of the robot's initial conditions.
Because the controller also knows the initial condition of the system, incorrect application of FDIA to the initial conditions would lead to detection.
The attacker is able to find the constant attack parameters to avoid detection by using their knowledge of the initial conditions.
The attacker was also assumed to have knowledge of the structure of the Jacobian matrix $\bm{J}$ used for the operation of the robot.
Note that if the structure of $\bm{J}$ is permuted, then the attack matrices should be permuted accordingly. On the other hand, the attacker was not required to know the geometries of the mobile robot or its mechanical details, such as wheel size, tread length, inertia, chassis material, and center of gravity.
Furthermore, the attacker was not required to know any details implemented in the controller, such as the tracking controller type or gains.

In the experiments, the FDIA's computational load was negligible compared to that of the controller and could be applied without affecting the real-time control capability.
For this experiment, the mobile robot was connected to the computer via an Ethernet cable to minimize time delay.

\subsection{Attack scenarios and results}

Two attack scenarios are shown as illustrative examples of how might the proposed FDIA can affect actual systems such as mobile robots.
The attack parameters $\bm{S}_u$, $\bm{S}_x$, and $\bm{d}_x$ are determined as presented in Section \ref{FDIAsolutions}.
The reflection attack (Scenarios 1 and 3) and scaling attack (Scenarios 2) are implemented with the attack matrices.
$\bm{d}_x$ is chosen to be $(\bm{I}_3-\bm{S}_x) \bm{p}_c(0)$ according to Condition 1 presented in Proposition 1.
The initial conditions are set to be $\bm{p}_c(0) = [0, 0.02, 0]^T$ with a zero initial orientation (0 degree) for the normal operation, Scenario 1 and Scenario 2. For Scenario 3, $\bm{p}_c(0) = [0, 0.02, \pi/6]^T$ with a non-zero initial orientation (30 degrees) are set to highlight the reflection about the $\theta_c(0)$ line. 
The attack parameters in these scenarios are determined as follows: 
\begin{itemize}
    \item {\bf Normal operation} (no attack):  \\$\bm{S}_x=\bm{I}_3, \bm{d}_x=\bm{0}, \bm{S}_u=\bm{I}_2,\bm{d}_u=\bm{0}$.
    \item  {\bf Scenario 1 - Reflection attack ($\beta_{11}=1, \theta_c(0)=0$)}: \\ \\$\bm{S}_x=\begin{bmatrix}1 & 0 & 0 \\ 0 & -1 & 0 \\0 & 0 & -1 \\\end{bmatrix}$, 
    $\bm{d}_x=\begin{bmatrix}0 \\ 0.04 \\ 0 \\\end{bmatrix}
    , \bm{S}_u=\begin{bmatrix}1 & 0 \\0 & -1\end{bmatrix}$, $\bm{d}_u= \bm{0}$.
    \item {\bf Scenario 2 - Scaling attack ($\beta_{11}=0.5$)}: \\ \\$\bm{S}_x=\begin{bmatrix}2 & 0 & 0 \\ 0 & 2 & 0 \\0 & 0 & 1 \\\end{bmatrix}, 
    \bm{d}_x=\begin{bmatrix}0 \\ -0.02 \\ 0 \\\end{bmatrix}$, $ \bm{S}_u=\begin{bmatrix}0.5 & 0 \\0 & 1 \\\end{bmatrix}$, \\$\bm{d}_u=\bm{0}$.
    \item  {\bf Scenario 3 - Reflection attack with non-zero initial orientation angle} ($\beta_{11}=1, \theta_c(0)=\pi/6$): \\ \\$\bm{S}_x=\begin{bmatrix} 0.5 & \sqrt{3}/2 & 0 \\ \sqrt{3}/2 & -0.5 & 0 \\0 & 0 & -1 \\\end{bmatrix}$, 
    $\bm{d}_x=\begin{bmatrix}-0.01\sqrt{3} \\ 0.03 \\ \pi/3 \\\end{bmatrix}
    , \bm{S}_u=\begin{bmatrix}1 & 0 \\0 & -1\end{bmatrix}$, $\bm{d}_u= \bm{0}$.
\end{itemize}

\begin{figure}[t]
	\centering
	\includegraphics[width=\columnwidth]{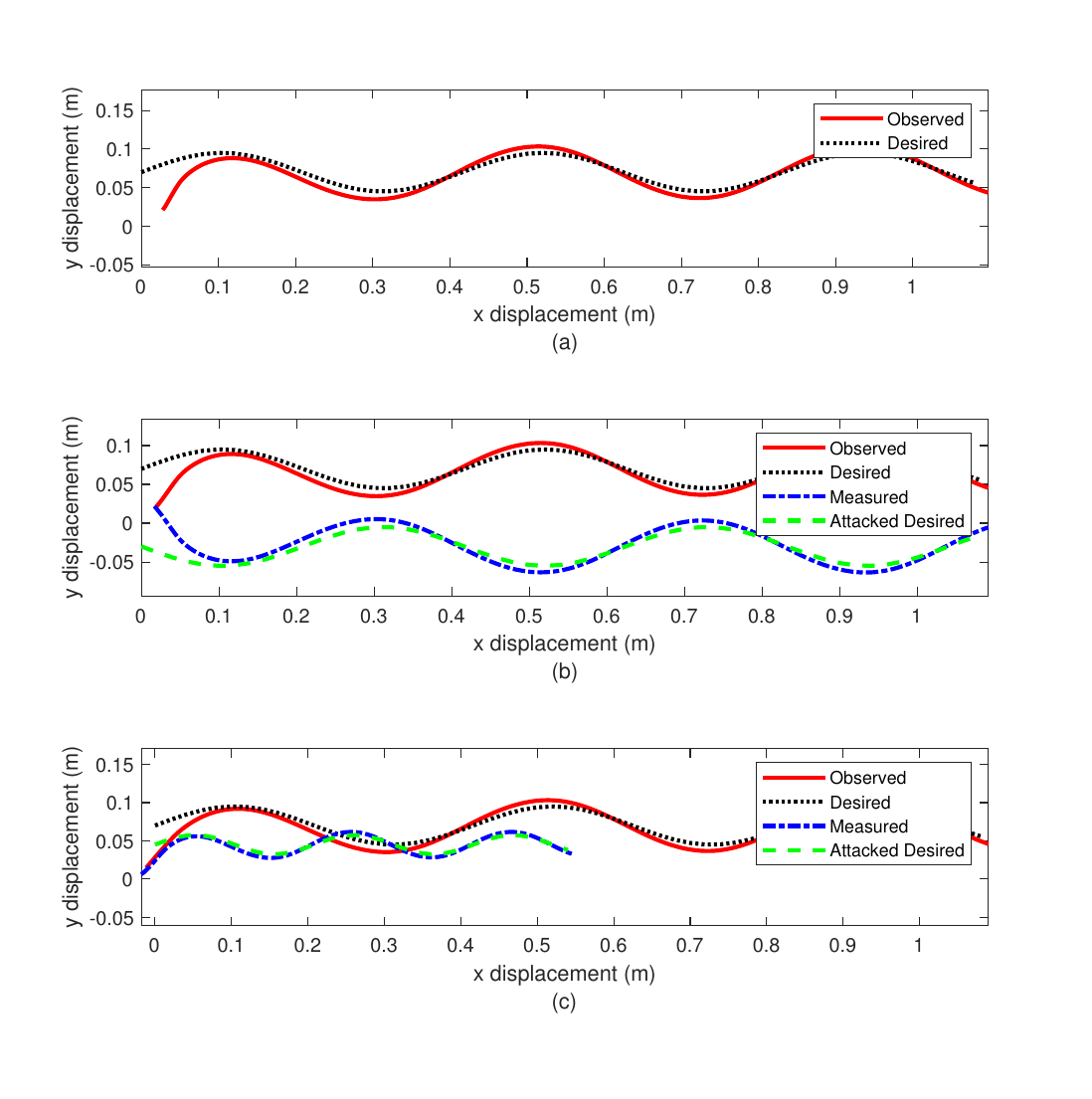}
	\caption{Robot position compared to position perceived by controller. Controller observation is identical to Original trajectory (a), but actual measured position of the robot follows a modified desired path in the reflection (scenario 1)(b) and scaling (scenario 2) (c) attack scenarios.}
	\label{desired_traj}
\end{figure}

\begin{figure}[t]
	\centering
	\includegraphics[width=1.0\columnwidth]{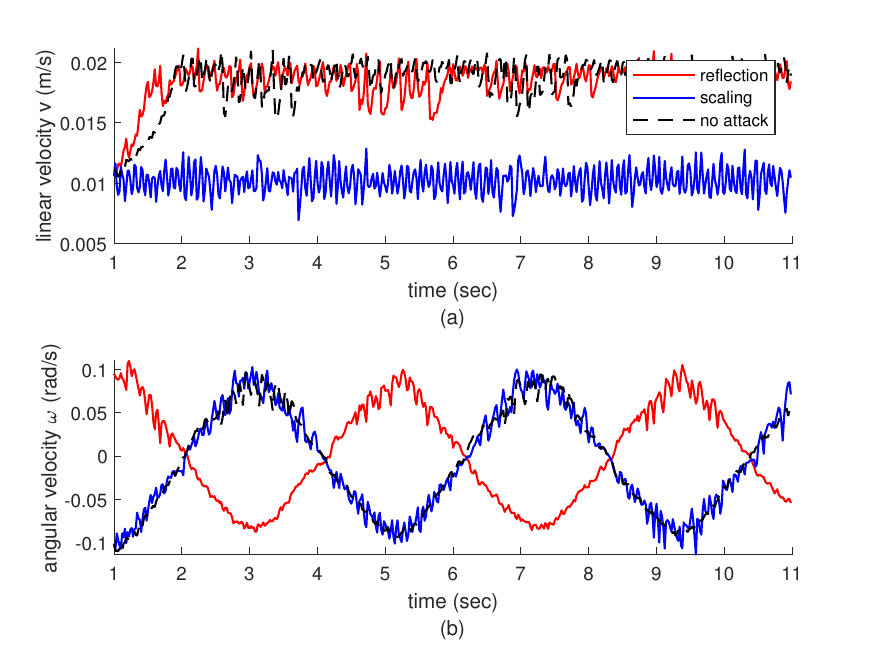}
	\caption{Control commands received by mobile robot in scenarios 1 and 2. (a) shows linear velocity command affected in the scaled attack, and (b) shows the angular velocity affected after the reflection attack.}
	\label{command_reflection_plot}
\end{figure}

\begin{figure}[t]
	\centering
	\includegraphics[width=1.0\columnwidth]{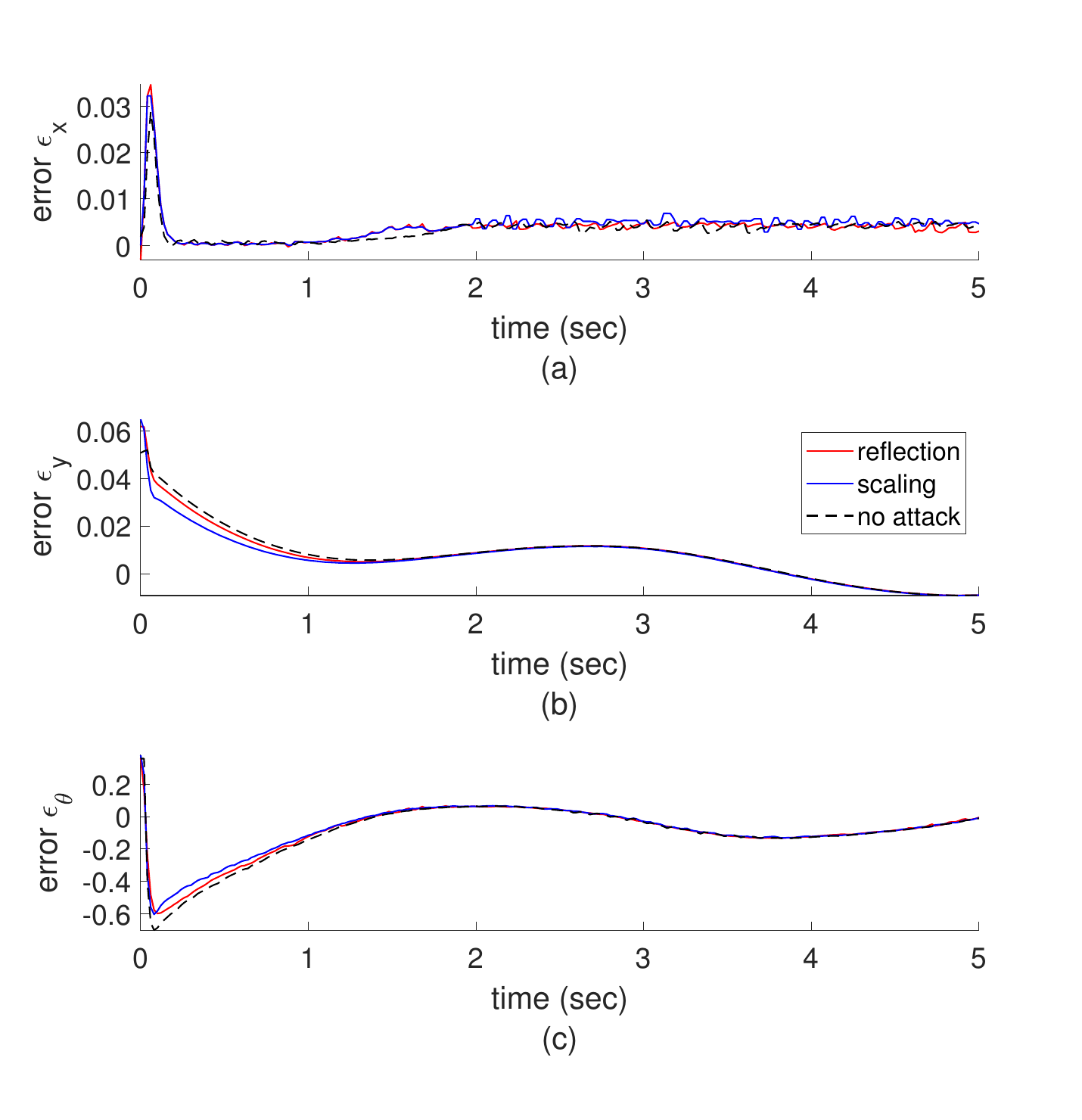}
	\caption{Error dynamics as observed by the controller for scenarios 1 and 2. Sub-figures (a), (b), and (c) show the $x$, $y$, and $\theta$ error used in the controller respectively. The overall error dynamics stayed the same regardless of the attack.}
	\label{error_reflection_plot}
\end{figure}

\begin{figure}[t]
	\centering
	\includegraphics[width=1\columnwidth]{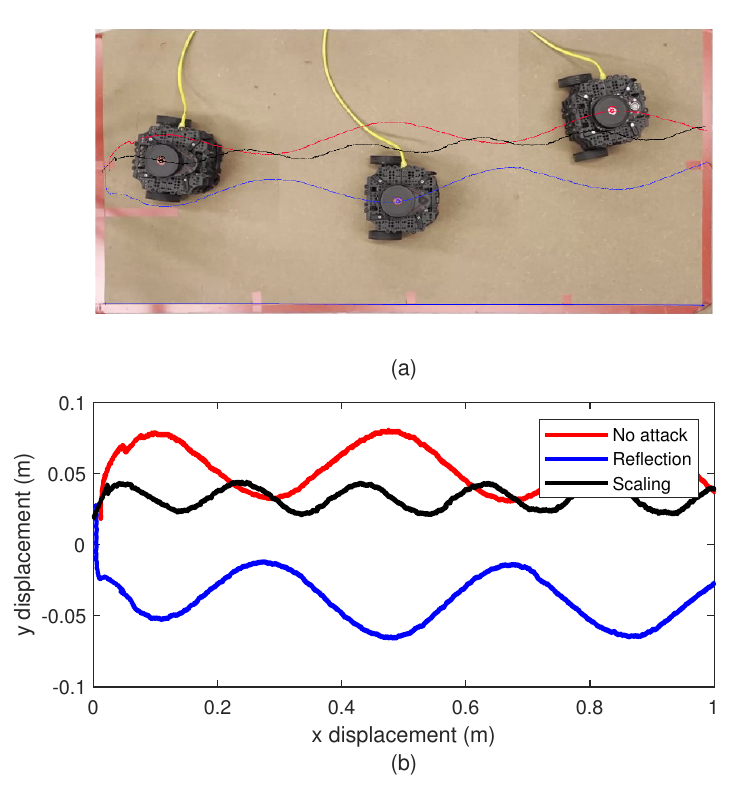}
	\caption{Trajectory of robot acquired through video analysis (scenarios 1 and 2). (a) Overlaid experimental screenshots. (b) Acquired robot trajectories.}
	\label{camera_traj}
\end{figure}
\begin{figure}[t]
	\centering
	\includegraphics[width=1\columnwidth]{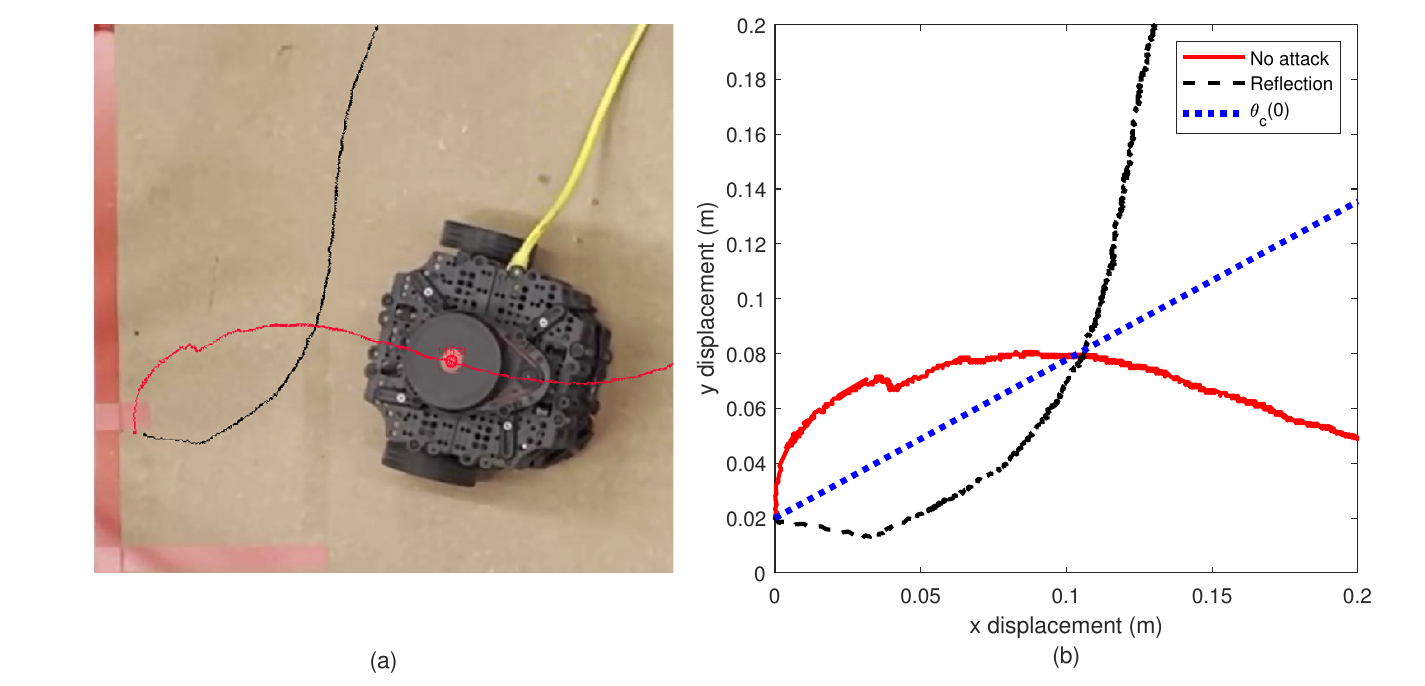}
	\caption{Reflection about non-zero initial orientation of $\pi/6$ rad (scenario 3): (a) overhead video and (b) video analysis with orientation shown in blue.}
	\label{reflection_traj}
\end{figure}
Three trials of the robot’s tracking of an identical desired trajectory under each FDIA attack scenario are reported. Fig. \ref{desired_traj} shows the desired trajectory affected by Scenario 1 and Scenario 2 for comparison. 
Fig. \ref{command_reflection_plot} shows different control commands received by the robot as it completes a sinusoidal path.
When not under any attack, linear velocity converged to around 0.02 m/s, and angular velocity showed a sinusoidal pattern with a period of 4 seconds.
The red and blue lines respectively depict the commands received by the robot under attack Scenario 1 (reflection) and Scenario 2 (scaling).

In Scenario 1, the linear velocity command showed the same tendency as the base case, while the angular velocity command was reflected.
In contrast, angular velocity command remained in phase with the base case in Scenario 2, but linear velocity converged to 0.01 m/s, following the scaling factor chosen beforehand.
This change in input commands resulted in reflection and scaling of the actual trajectory of the robot shown in Fig. \ref{desired_traj}.

The perfectly undetectable attack is carried out in the feedback loop according to the above scenarios, leading to observed positions that do not reflect the robot’s current state.
Fig. \ref{desired_traj} also shows the comparison between the actual trajectory and the observed position perceived by the separate controller. 
After the attack is applied, the observed position of the robot matches very well with the desired trajectory as in Fig. \ref{desired_traj}.
The controller observed that the robot is converging well to the predefined desired trajectory in black.
This observation aligns with Fig. \ref{error_reflection_plot}, which shows the error dynamics observed by the controller.
However, the robot's actual position is at the measured position in blue, which follows the attacked desired trajectory in green. Even when system behavior is significantly changed through FDIA, there are no apparent signs of such deviation from the intended trajectory in the error dynamics to characterize the application of such an attack.
Successful application of a residual-based detection method \cite{Sandberg22} is unlikely in the attack scenarios presented.
Fig. \ref{camera_traj} shows the measured position of the robot in each scenario as observed from an overhead position.

In addition, Fig.\ref{reflection_traj} shows Scenario 3 (reflection attack with a non-zero orientation angle) for a clearer visual representation of the reflection attack.
The robot's initial condition $\bm{p}_c(0)[\,0,\, 0.02,\, \pi/6\,]^T$ was used. A nonzero initial orientation $\theta_c(0)=\pi/6$ sets the axis of reflection as illustrated in Fig. \ref{mobilerobot_reflectionscaling} (a).

\section{Perfectly undetectable FDIA resilient state monitoring}
\label{statemonitoring}
\subsection{Affine transformation resilient state monitoring signature functions}

Based on the assumption of affine transformation-based perfectly undetectable FDIA described above, the presence of non-trivial attack matrices $\bm{S}_x$, $\bm{d}_x$, $\bm{S}_u$, and $\bm{d}_u$ that realize perfectly undetectable FDIAs has been demonstrated.  {\bf Proposition 1} is a very strong condition that makes it theoretically impossible for the controller to detect an attack based on the observation of compromised observables $\tilde{\bm{p}}_c$ corresponding to command $\bm{u}$.  

\begin{figure}[t]
	\centering
	\includegraphics[width=1.0\columnwidth]{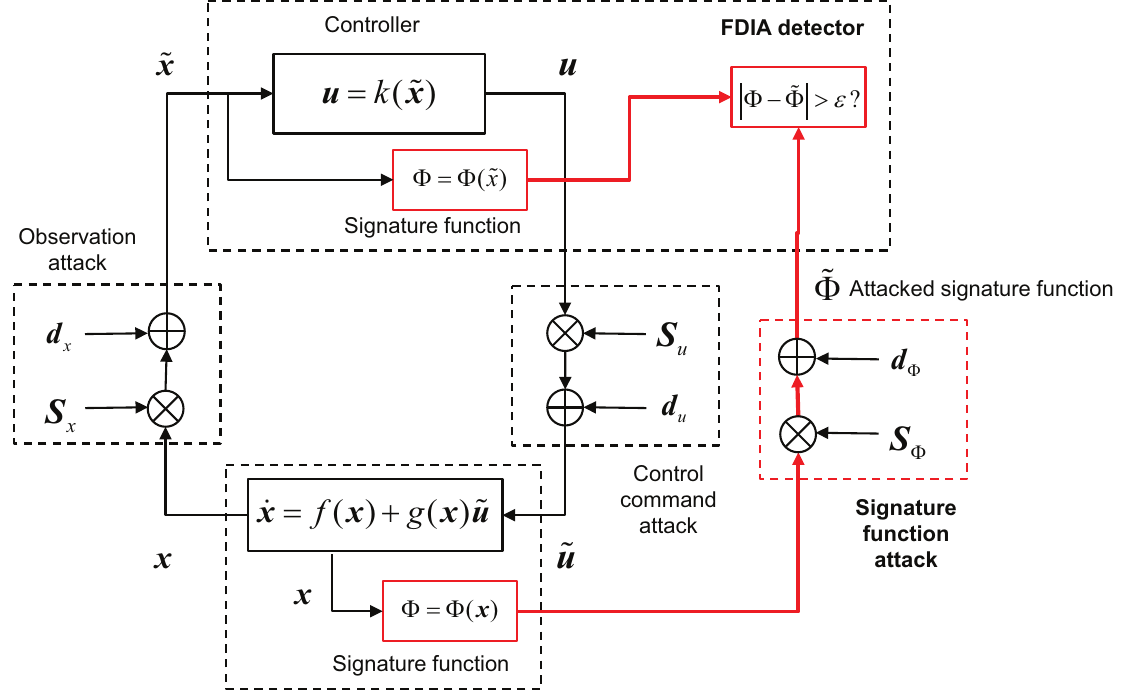}
	\caption{Continuous state monitoring by using a signature function under affine transformation based FDIA}
	\label{monitoringfunction}
\end{figure}

The proposed countermeasure is to implement a separate function $\Phi(\bm{x})$ for state monitoring in the plant, evaluated based on the ground truth states, and compare its counterpart evaluated in the controller, as illustrated in Fig. \ref{monitoringfunction}. Any discrepancies between them that exceed an acceptable level of noise could indicate a possible attack. In the literature, several methods to detect FDIAs have been proposed, e.g., \cite{Sandberg22,Pasqualetti13,hoehn2016detection}. It should be noted that, in contrast to conventional studies in the literature, this work assumes that the communication channel transmitting the output of the signature function is also susceptible to affine transformation FDIA with attack matrices $\bm{S}_{\Phi}$ and $\bm{d}_{\Phi}$. As shown in Fig. \ref{monitoringfunction}, the attacker might determine $\bm{S}_{\Phi}$ and $\bm{d}_{\Phi}$ based on the eavesdropping of $\bm{x}$. 

The proposed state monitoring signature function (SMSF) serves as an authentication method similar to hash functions \cite{Merkle1989} and auxiliary systems \cite{Schellenberger}.
In contrast with a hash function, the SMSF can be tailored to suit the control system under operation; for instance an SMSF can be formulated to accommodate varying state dimensions.
The output of an SMSF can be designed to be smooth, unlike that of a hash function.
This smoothness allows the SMSF to be more interpretable along a smooth trajectory in the presence of noise.
The SMSF is also a static function that does not require stabilization, contrary to dynamic auxiliary systems.
The static design enhances resilience against certain attacks and simplifies implementation.
These features collectively create a robust mechanism for detecting FDIAs.

The proposed SMSF is contracted to be resilient to both scaling and reflection FDIAs as described in Appendix \ref{appen_attacksincos}.

{\bf Proposition 4. Scaling and reflection attack resilient SMSF.}
As analyzed below in detail, the SMSF must be injective, nonlinear, and noninvertible. The noninvertiblility may be achieved by choosing a dimensional reduction function, such as a scalar function that takes multiple inputs. Suppose a scalar signature function $\Phi(\bm{x})$ of the state $\bm{x}$. 
If 
$\Phi \left( {{{\bm{S}}_x}{\bm{x}} + {{\bm{d}}_x}} \right) = {{\bm{S}}_\Phi }\Phi \left( {\bm{x}} \right) + {{\bm{d}}_\Phi }$
holds only when 
${{\bm{S}}_\Phi } =1,{{\bm{d}}_\Phi } = 0$, the function  $\Phi ( \bm{x})$ is appropriate as an FDIA resilient SMSF at least by affine transforms. 

{\bf Remark 5: Inappropriateness of linear functions for attack-resilient state monitoring.}
Note that a linear function is not appropriate at all including the integration of inputs (total control effort), as the input-to-output relationship is linear, therefore, a scaling attack ($\beta$ and $1/\beta$ combination, see Appendix \ref{appen_attacksincos} linear functions) is always applicable. Also a linear signature function may be easily estimated by standard least squares estimation techniques, necessitating that the function be nonlinear to resist FDIA.

\subsection{Construction of signature functions for continuous state monitoring}
\label{constructionofsignaturefunctions}

Consider a positive definite function as a candidate signature function:
$
\Phi ({\bm{x}}) > 0,{\bm{x}} \ne 0,
\Phi ({\bm{x}}) = 0,{\bm{x}} = 0,
{\bm{x}} \subset S \in {R^n}
$
where $S$ is an operational range of $\bm{x}$.  This non-negative property enforces the additive attack $\bm{d}_\Phi=0$ otherwise it is detectable. Note that $\bm{x}(0)\neq \bm{0}$ may be used as long as the positive definiteness is achieved.  
{\it Proof:}
$\Phi ( {\bm{x}}  ) = \bm{S}_\Phi\Phi ({\bm{x}}) + \bm{d}_\Phi= 0$ holds
if and only if
$\bm{d}_{\Phi} = 0$. 
$\blacksquare$

Although it is a nonlinear function, a polynomial function
$\Phi (x,y) = {x^2} + {y^2}$ 
is not appropriate since the function is known to have the {\it homogeneity of degree two}, i.e., 
$\Phi (\alpha x, \alpha y) = {\alpha^2}{x^2} + {\alpha^2}{y^2} = {\alpha^2}\Phi (x,y)$. The attacker can introduce a FDIA that multiplies the output of the signature function by the square of the scaling factor, rendering the attack undetectable. 
Also, this candidate function only concerns the distance from the origin $\sqrt {{x^2} + {y^2}}$ exhibits radial symmetry about the origin and therefore deemed inappropriate. 

Without loss of generality, we can consider constructing a monitoring signature that is a) positive definite (taking 0 only at the origin) to detect linear translation attacks by $\bm{d}_\Phi$, b) non-invariant under scaling and lacking symmetry to detect scaling and reflections by $\bm{S}_\Phi$. Although guaranteeing the detection of general affine transformations is challenging, roughly speaking, c) having asymmetric contours (or level sets) would be necessary. Below, a polynomial function of the state variables is considered as a candidate signature function, constructed according to the following design guidelines.

\begin{enumerate}
    \item Use even-powered terms (with at least two different powers) to ensure non-negativity and avoid homogeneity of any specific degree. Unlike Lyapunov or ``Lyapunov-like" functions used in the literature, the negative (semi) definiteness of the derivative of the function is not strictly necessary for monitoring purposes. 
    \item Incorporate odd-powered terms to ensure that sign flips (e.g., $x^3$ and $\left(-x\right)^3$) compute differently. Note that odd-powered terms must be introduced as a part of even-powered terms to ensure non-negativity. 
    \item  Include coupled terms between different variables to introduce asymmetry (e.g., $x^2y$). Note that these asymmetries must be sufficiently nonlinear to prevent reversal through a linear transformation.
\end{enumerate}

Consider a candidate signature function that involves two variables $x$ and $y$ and extends up to the quartic degree. According to Requirement 1, $x^2, x^4, y^2, y^4$ must be included. According to Requirements 2 and 3, some (or all) terms like $x^3, y^3, xy^2, x^2y$ should be included. For those reasons, functions of only up to the quadratic degree are not suitable.
Note that one of the state variables $\theta$ is not used for simplicity. While $\Phi$ is positive semi-definite in this case, because the attacker is not able to implement a pure rotation attack about the initial position along a trajectory, this property does not impact the ability of state monitoring.  

{\bf Remark 6: Constructing more complex signature functions.} To enhance resilience against estimation attacks, one can consider incorporating a variety of mathematical constructs beyond simple polynomials. These can include exponential functions for rapid non-linear growth, trigonometric functions for periodic behavior, discontinuous functions for abrupt changes, composite functions combining different types, piecewise functions with region-specific behaviors, and recursive functions for iterative complexity. When constructing such functions, we should balance complexity with computational efficiency. 

\begin{figure}[t]
	\centering
	\includegraphics[width=1.0\columnwidth]{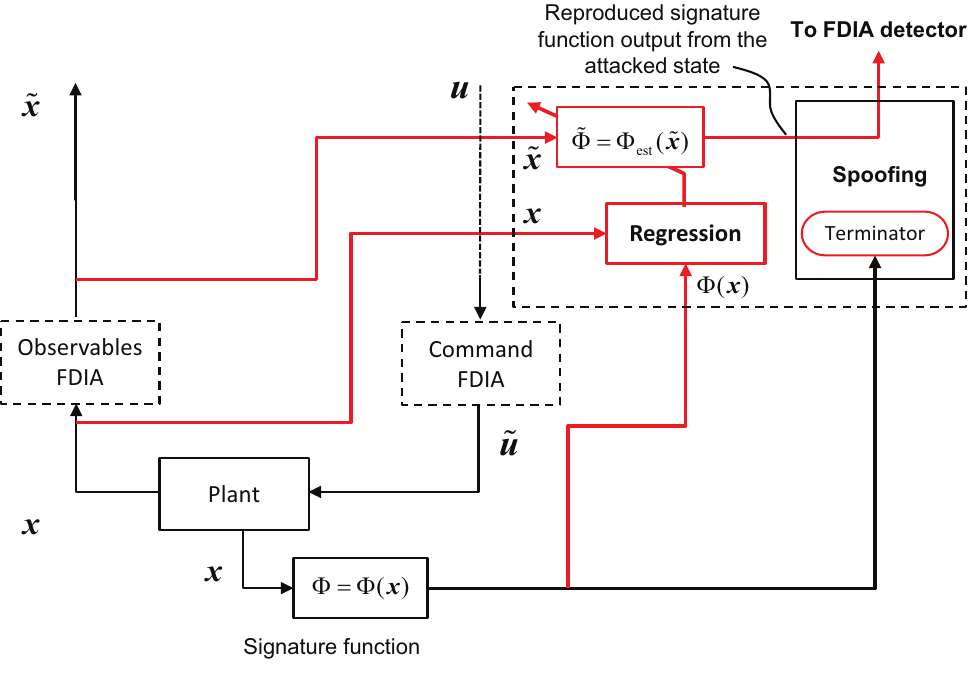}
	\caption{Spoofing attack to state monitoring via adversarial regression of signature function}
	\label{monitoringfunction_spoofing}
\end{figure}

As another attack scenario, an intelligent attacker can estimate the signature function through regression to fully reproduce $\Phi(x)$ and completely alter the signal to align with $\hat{\Phi}$, as shown in Fig. \ref{monitoringfunction_spoofing}, making the attack remain undetectable.
This implies that the SMSF remains secure only until an attacker, who can intercept the state variables $\bm{x}$ and $\Phi(\bm{x})$, fully estimates the function, and the security may be quantified by its sampling complexity. The state monitoring approach will lose all effectiveness immediately if $\bm{\Phi}(\bm{x})$ is fully known by the attacker. 

When implementing the SMSF approach, it is assumed that the structure of the function, such as being a polynomial function of the state variables and its degree, may be known. However, the coefficients are not known at the beginning of system operation. It is safe to change the coefficients before each system operation as a 'moving target'. However, the coefficients remain fixed during each operation, as changing them would require additional communication between the controller and plant that may be intercepted.

$\Phi(x,y)$ may be estimated by using polynomial regression (PR), Gaussian Process Regression (GPR), Neural Network (NN) regression or other alternatives. 
The VC dimension \cite{Blumer1989} of $\Phi(x,y)$ is a reasonable starting point for estimation. However, this should be considered a minimum guideline; more samples may be beneficial for robust estimation, particularly in complex regions of the function domain. The sample complexity \cite{Meyer2023,Teranishi2023}, which quantifies the number of examples needed to learn a function to a given accuracy, increases with the VC dimension and the desired precision of the estimate. In practice, the required number of samples can be significantly higher than this lower bound, especially for complex, nonlinear functions. As demonstrated in the illustrative example below, the estimation of $\Phi(x,y)$ only from intercepted samples along the realized trajectory is much more challenging for the attacker, further increasing the effective sample complexity.

{\bf Definition 2: Security of state monitoring along a trajectory against adversarial estimation}
Let $\hat{\Phi}({\bm{x}})$ be the function estimated by the attacker using $N$ intercepted samples ${{\bm{x}}(t_i)}, i=1, \cdots, N$ collected along the trajectory from $t=0$ to the current time, where $0 \leq t_1 < t_2 < ... < t_N$. 
The security of the SMSF is maintained if:
\[
\sup_{{\bm{x}} \in S} ||\bm{\Phi}({\bm{x}}) - \hat{\bm{\Phi}}({\bm{x}})|| > \epsilon > 0
\]
where $S \subset \mathbb{R}^n$ is the relevant state space. If this condition holds, the attacker cannot alter $\bm{\Phi}({\bm{x}})$ being sent from the plant to the controller below a threshold $\epsilon$ throughout the state space, and therefore any attack can be detected by the controller.

{\bf Remark 7.}
As an additional security measure, it is recommended to encrypt $\Phi$ and evaluate it using methods such as homomorphic encryption applicable to real-time control \cite{RN60}. It is expected that the security of the signature function improves in accordance with the sample complexity in both the cryptosystem and the target plant dynamic model \cite{Teranishi2023}. However, note that the application of encryption does not fully prevent the risk of signature function estimation.

\subsection{Illustrative example: adversarial estimation scenarios}

\begin{figure}[t]
	\centering
	\includegraphics[width=1\columnwidth]{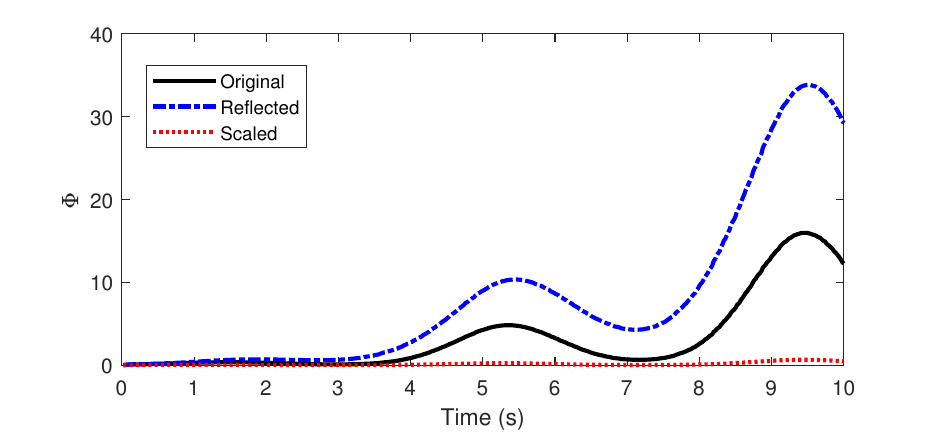}
	\caption{State monitoring by using an example signature function $\tilde{\Phi} (x,y)$ evaluated at the plant under perfectly undetectable attacks: Scenarios 1 (reflection) and 2 (scaling) compared to $\Phi(\tilde{x}, \tilde{y})$ evaluated at the controller.}
	\label{monitoringfunction_plot}
\end{figure}

A quartic, scalar signature function can be constructed according to the guidelines in Section \ref{constructionofsignaturefunctions} as:
\begin{eqnarray}
    && \!\!\!\!\!\!\!\! \Phi (x,y) = {x^4} + {y^4} + {(x - 50xy)^2} + {(xy - 5y)^2} \nonumber \\
    && \!\!\!\!\!\!\!\!\!\!\!\!\!\!\!\!\!\!\!  = {x^4} + {y^4} + {x^2} + 25{y^2} - 100{x^2}y - 10x{y^2} + 2501{x^2}{y^2}  \label{representativemonitoringfunction}
\end{eqnarray}
for which attack parameters $S_\Phi$ and $d_\Phi$ such that $S_\Phi \Phi(x) + d_\Phi = \Phi(\tilde x)$ do not exist, and the affine transformation attack on the signature function shown in Fig. \ref{monitoringfunction} is impossible. 
With this signature function, under the perfectly undetectable attacks {\bf Scenario 1 and 2} in Section \ref{experiments}, for example, when the communication line is not compromised, i.e., 
$S_\Phi =1, d_\Phi=0$, $\Phi(\tilde{\bm{x}})$ evaluated at the controller and $\tilde{\Phi}(\bm{x})$ evaluated at the plant along the attacked trajectories are shown in Fig. \ref{monitoringfunction_plot}.
In this specific case, the scaling attack affects the $\Phi$ values more drastically as the magnitude of the robot’s position sees greater changes than those of the reflection case, leading to the detection of the attacks. 
As a result, a significant disparity is indicative of the attack on the system, and only when no attack is performed, $\Phi(\tilde{\bm{x}})=\tilde{\Phi}(\bm{x})$ holds. 

\begin{figure}[t]
	\centering
	\includegraphics[width=1\columnwidth]{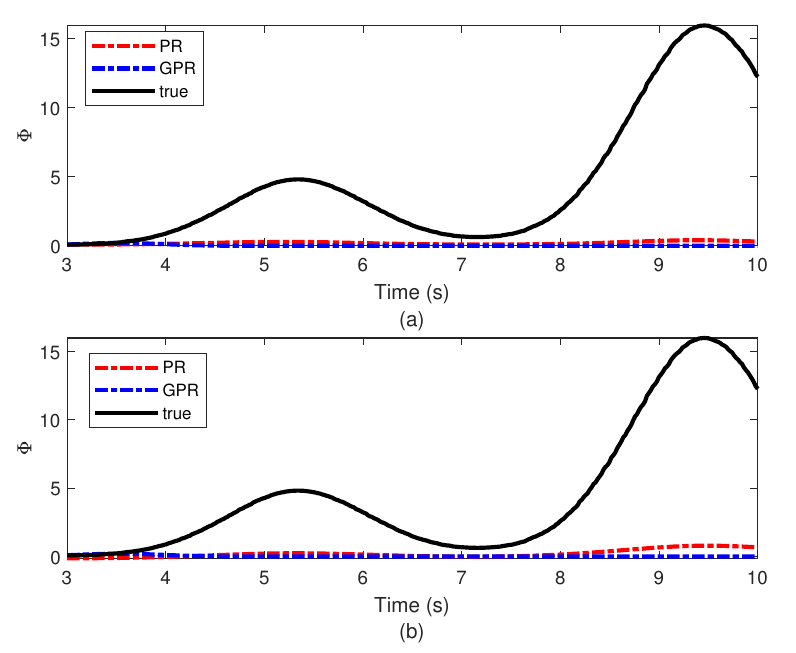}
 \caption{
 Unsuccessful adversarial estimation $\hat{\Phi}(\tilde{x})$ with 150 samples for (a) Scenario 1 (reflection) and (b) Scenario 2 (scaling).}
	\label{exp_metrics}
\end{figure}

Next, as an alternative strategy for the attacker, they attempt to directly estimate $\Phi$ through collected data.  
The complexity of such adversarial estimation for the example signature function given in (\ref{representativemonitoringfunction}) is evaluated along the trajectories demonstrated in {\bf Scenarios 1 and 2} in Section \ref{experiments}. As representative regression techniques, polynomial regression (PR) and Gaussian process regression (GPR), are applied. For PR, it is plausible to assume that the attacker knows the degree of the polynomial, and thus all the polynomial bases are known. The estimation is then performed to identify the coefficients. 
The main challenge for the attacker lies in their ability to estimate $\Phi$ (denoted as $\hat{\Phi}$) by intercepting the attacked trajectory $\bm{x}$, with sufficient accuracy to  reproduce $\Phi(\tilde{x})$. The success of this adversarial estimation and reproduction is critical, as it makes the state monitoring process based on $\Phi$ fully ineffective.
Since the VC dimension of (\ref{representativemonitoringfunction}) is 15, seeing if using 150 sample points (15 $\times$ 10) to for 3 seconds, following the heuristic of using about 10 times the VC dimension. 
Here, the attacker is assumed to eavesdrop for the first 3 seconds ($t=0-3$) for each of the operations, and then uses the acquired $\hat \Phi$ to predict the following values. 
In this instance the attacker allows the controller to monitor $\Phi(\tilde{x})$ till $t=3$ 
In this instance, the attacker used a sample limited to the trajectory, leading to poorer estimations. 
Fig. \ref{exp_metrics} shows the comparison between $\hat\Phi(\tilde x)$ and $\Phi(\tilde x)$ over time. 
As the sample is not distributed evenly within the workspace, only along the realized trajectories, the estimation fails, and the security of the monitoring approach is maintained.

\begin{figure}[t]
	\centering
	\includegraphics[width=1\columnwidth]{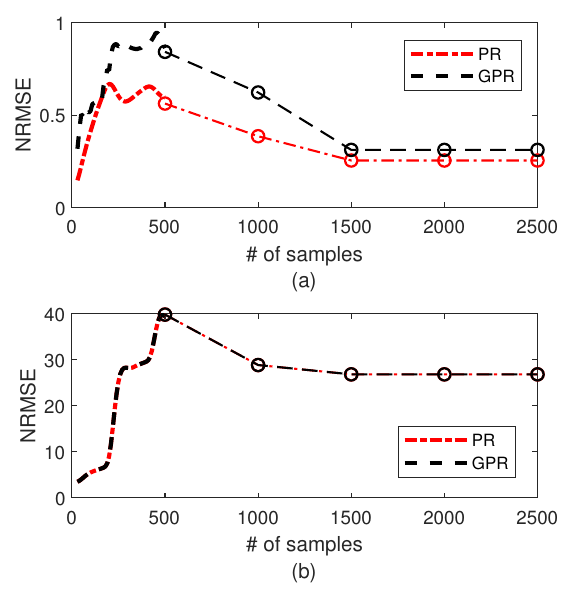}
 \caption{Performance of signature function estimation: Normalized RMSE with respect to the number of samples $N$ with experimental and simulated data. (a) for Scenario 1 (reflection) and (b) Scenarios 2 (scaling). Simulated data displayed for points beyond 500 samples.}
	\label{sim_metrics}
\end{figure}

\begin{figure}[t]
	\centering
	\includegraphics[width=1\columnwidth]{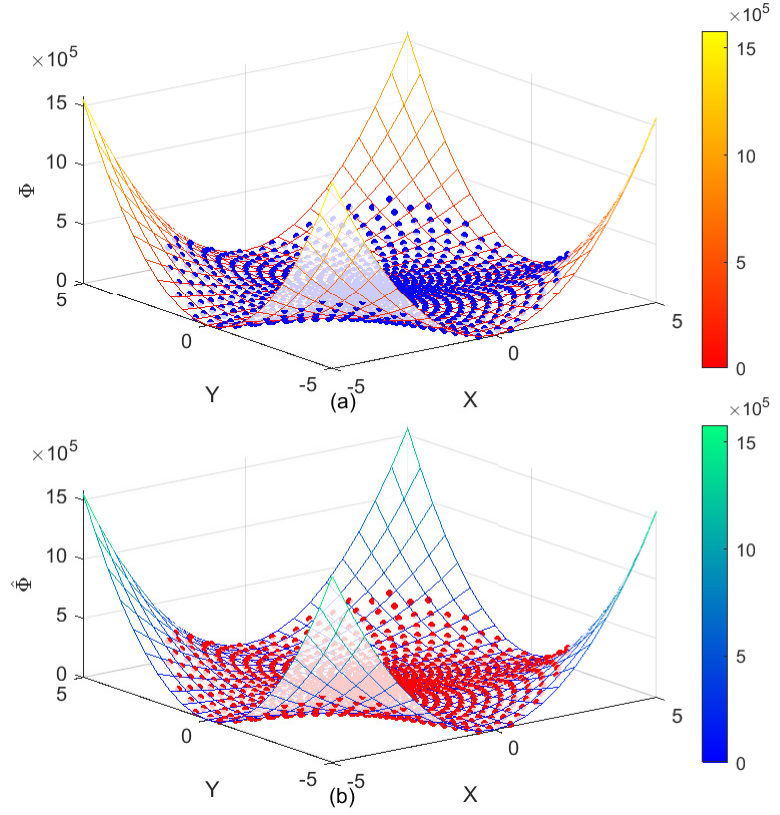}
	\caption{Polynomial regression over a large number of samples $N$=1000 with a simulated spiral trajectory with a wide coverage of the workspace. (a) True $\Phi$ and (b) estimated $\hat{\Phi}$ with a NRMSE value of 0.059.}
	\label{fixed_RMSE}
\end{figure}

From the attacker's perspective, there may be two potential attempts to improve the performance of $\hat\Phi$: 1) increased number of samples, and 2) a trajectory that provides a wider coverage of the workspace. 
This allows for a bigger window for the controller to detect the presence of an FDIA before $\hat\Phi$ can be estimated. 
Fig. \ref{sim_metrics} shows the performance of the estimation with increased numbers of samples for regression. 
For the first 500 samples, the experimental data from Section \ref{experiments} was used, followed by the data generated by simulation. 
Normalized root mean square error (NRMSE) is shown as the metric of fitness. 
In general, increasing the number of samples improved the performance of estimation. Nevertheless, neither of the attack scenarios provides perfect estimation, probably due to insufficient coverage of the workspace along the attacked trajectories. 
If the user operates the robot along a trajectory that widely explores the workspace, the attacker would collect a sufficiently rich dataset. Fig. \ref{fixed_RMSE} shows the performance of polynomial regression performed with data from a fictitious spiraling trajectory simulated up to 1000 samples. With a sample size of 150, the recommended sample size based on the VC dimension, the estimation result was significantly better compared to the previously considered sinusoidal trajectory with a NRMSE of only 0.26. The distribution of samples was observed to be the main factor for successful estimation. It is understandable that estimation of the signature function is eventually achieved by the attacker, indicating the necessity of frequent updates of the function.

\section{Discussion}
\label{discussion}

Identification and classification of potential cyberattacks are important tasks when working with networked robots, as a thorough understanding of the effects and conditions then allows for further studies on defense strategies. Preexisting discussions of deceptive and undetectable attacks are often limited, with strong assumptions restraining the attacker's capability and focusing on simple systems \cite{SMITH201190,Schellenberger}. However, even without conditions such as full knowledge of system dynamics or linearity, substantial attacks could be applied to systems.
 
The mobile robot tracking control experiment in this paper demonstrates that for an attacker capable of compromising both legs of a networked control system, the only information required by the attacker was the structure of the Jacobian $\bm{J}$ and initial conditions, unlike covert attacks that necessitate perfect knowledge of the plant dynamics. In the studies by Zhai et al. \cite{zhai2022graph} and Sandberg et al. \cite{Sandberg22}, the realization of covert attacks is formulated as an optimization problem using the plant dynamic model. These studies allow for control-theoretic discussion and are thus of academic interest, but their implementation may be challenging from an engineering perspective.

In contrast, affine transformation seems to be a more practical attack method. Relatively simplistic attacks with constant attack parameters, which do not alter the degree of the closed-loop dynamics, were effective to significantly modify the robot's behavior while remaining undetectable from the controller's perspective. 
The affine transformation-based FDIAs are highly effective on multi-dimensional robotic systems rather than low-dimensional systems but with high degree. By manipulating variables of the same physical quantities through operations such as scaling or reflection, attackers can significantly alter a robot's behavior while maintaining mathematical consistency. This attack exploits the multi-dimensional nature of robotic systems, making attacks difficult or impossible to detect yet simple to implement, even with partial system knowledge. The preservation of mathematical structures in these multi-dimensional spaces poses significant challenges for traditional detection methods.

Regarding electronic watermarking \cite{du2021secure}, even if white noise is added at the controller, the observed dynamics remain unchanged as shown in (\ref{dotpprime}), so the effect of white noise will be accurately restored on the observation side. This means that the covariance of the estimation error by the observer does not change, making existing watermarking methods ineffective.

Note that the existence of perfectly undetectable FDIA is guaranteed for linear plants \cite{ueda2024affine} but not necessarily for all nonlinear dynamic systems. While Proposition 1 provides a general framework for nonlinear dynamics in affine form, the existence of solutions depends on the specific system characteristics. This paper demonstrates that such attacks exist for mobile robot dynamics, a significant finding in robotic security. However, fully generalizing perfectly undetectable FDIA to all nonlinear systems requires further research, as conditions for their existence may vary widely across different nonlinear plants.

As a countermeasure against perfectly undetectable FDIAs, a state monitoring approach using a signature function has been proposed. A sufficiently complex polynomial function is resilient against affine transformations. However, it should be noted that this approach has partial vulnerability to estimation attacks; given enough time and data, an attacker could potentially estimate the signature function through regression techniques, especially if they can observe a trajectory that covers a wide range of the workspace. Additionally, the example signature function discussed and analyzed in this paper is merely satisfactory. The signature function and its update frequency could be optimally determined once intended trajectories are given.

\section{Conclusion}
\label{conclusion}
The paper focuses on a mobile robot trajectory tracking control system as a case study, highlighting the susceptibility of nonlinear systems with partially linear dynamic properties and symmetries to this type of attack. The experimental results using a Turtlebot 3 platform validate the practicality of implementing such attacks, emphasizing the urgent need for more robust security measures in Cyber Physical Systems (CPS).
This paper demonstrated that a typical mobile robot trajectory tracking control system is susceptible to perfectly undetectable false data injection attacks. Two specific types of perfectly undetectable FDIA are possible: scaling and reflection attacks, both based on affine transformations. These findings demonstrate the critical need for more robust detection mechanisms and resilient control strategies to protect such systems. 

Future work will focus on developing effective countermeasures to mitigate the risks associated with these sophisticated cyberattacks and enhance system security in real-world applications including customization of SMSFs. Additionally, exploring response strategies that leverage machine learning could offer promising avenues for advancing the resilience of CPS against increasingly complex attack vectors. Furthermore, the introduction of time-variant perfectly undetectable attacks could lead to more sophisticated and powerful attacks compared to the simple scenarios mentioned in this paper.

\appendices

 \section{Perfectly undetectable FDIA from the plant's perspective}
\label{appen_perfectFDIAplant}

{\bf Definition A1: Perfectly undetectable FDIA from the plant's perspective} (Milosevic 2021 \cite{Sandberg22,GRACY21}). Let $y(x(0),u,a)$ denote the response of the system for the initial condition $x(0)$, input $u(t)$, and attack signal $a(t)$. The attack is perfectly undetectable if 
\begin{equation}
    y(x(0),u,a)=y(x(0),u,0), t \geq 0
    \label{perfectFDIAplantdef}
\end{equation}

The attacker leaves no traces in the measurements of $y$. Consequently, the attacker can impact the system's performance or behavior without being detected by an attack detector that utilizes $y$ for attack detection. Research has shown that (\ref{perfectFDIAplantdef}) can be achieved through zero dynamics attacks in the presence of transmission zeros \cite{hoehn2016detection}.

 \section{Linear FDIA vulnerability in polynomial and trigonometric functions}
\label{appen_attacksincos}

Consider a scalar function $g(x)$ with scalar attack parameters $\alpha (\neq0)$ affecting the output and $\beta (\neq0)$ affecting the input, resulting in the function $\tilde{g}(x)= \alpha g(\beta x)$.
Examine conditions where $g(x)=\tilde{g}(x)$ holds for all $x$.  The function is said to be susceptible to linear attacks if non-trivial solutions for $\alpha$ and $\beta$ exist other than the trivial case (i.e., $\alpha=\beta=1$). Results for representative functions and brief proofs are presented below.

{\bf Proposition B1: Linear FDIA vulnerability of representative scalar functions}.
\begin{enumerate}
\item $g(x)=c x$: $\alpha g(\beta c x)= \alpha \beta c x$. There are an infinite number of solutions that satisfy $\alpha \beta=1$ for such linear functions. Note that the attacker does not require knowledge of the coefficient $c$. 
\item $g(\theta)=\cos(\theta)$. Consider its first and second derivatives with respect to $\theta$: $g'(\theta)=
    -\sin (\theta) = -\alpha \beta \sin(\beta \theta)$, and $g''(\theta)=
   -\cos(\theta)=-\alpha \beta^2 \cos(\beta \theta)$. 
$\beta=\pm1$ since $\beta^2=1$. When $\beta=-1$ (non-trivial case), $\alpha=-1$ since $\alpha\beta=1$.
\item $g(\theta)=\sin(\theta)$:  Similar analysis to the above yields $\beta=-1$ (non-trivial case) and $\alpha=1$ since $\alpha\beta=-1$.
\item $g(x)=c x^2$: $\alpha g(\beta x)= \alpha c \beta^2 x^2$. There are an infinite number of solutions that satisfy $\alpha \beta^2=1$. Note that the attacker does not require knowledge of the coefficient $c$. 
\item $g(x)= c e^x$:  $\alpha g(\beta x)= \alpha c e^{\beta x}$. Comparing the first derivative functions with respect to $x$:  $g'(x)= c e^x =  \alpha c \beta e^{\beta x} $ yields only the trivial case,  $\alpha=\beta=1$. The exponential function is resistant to linear attacks. 
\end{enumerate}

\bibliographystyle{IEEEtran}
\bibliography{citations}

\begin{IEEEbiography}[{\includegraphics[width=1in,height=1.25in,clip,keepaspectratio]{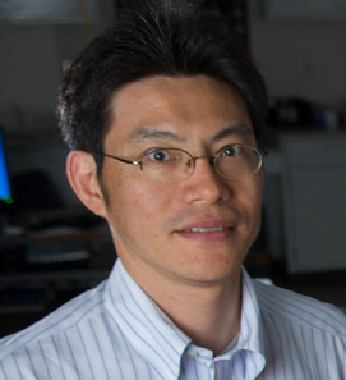}}]{Jun Ueda}(SM 2018) received the B.S., M.S., and Ph.D. degrees from Kyoto University, Japan, in 1994, 1996, and 2002 all in Mechanical Engineering. From 1996 to 2000, he was a Research Engineer at the Advanced Technology Research and Development Center, Mitsubishi Electric Corporation, Japan. He was an Assistant Professor of Nara Institute of Science and Technology, Japan, from 2002 to 2008. During 2005-2008, he was a visiting scholar and lecturer in the Department of Mechanical Engineering, Massachusetts Institute of Technology.  He joined the G. W. Woodruff School of Mechanical Engineering at the Georgia Institute of Technology as an Assistant Professor in 2008 where he is currently a Professor.  He served as the Director for the Robotics PhD Program at Georgia Tech for 2015-2017. He currently serves as a Senior Editor for IEEE/ASME Transactions on Mechatronics. He is the author of Cellular Actuators: Modularity and Variability in Muscle-Inspired Actuation, Butterworth-Heinemann, 2017. He received Fanuc FA Robot Foundation Best Paper Award in 2005, IEEE Robotics and Automation Society Early Academic Career Award in 2009, Advanced Robotics Best Paper Award in 2015, and Nagamori Award in 2021. 
\end{IEEEbiography}

\begin{IEEEbiography}[{\includegraphics[width=1in,height=1.25in,clip,keepaspectratio]{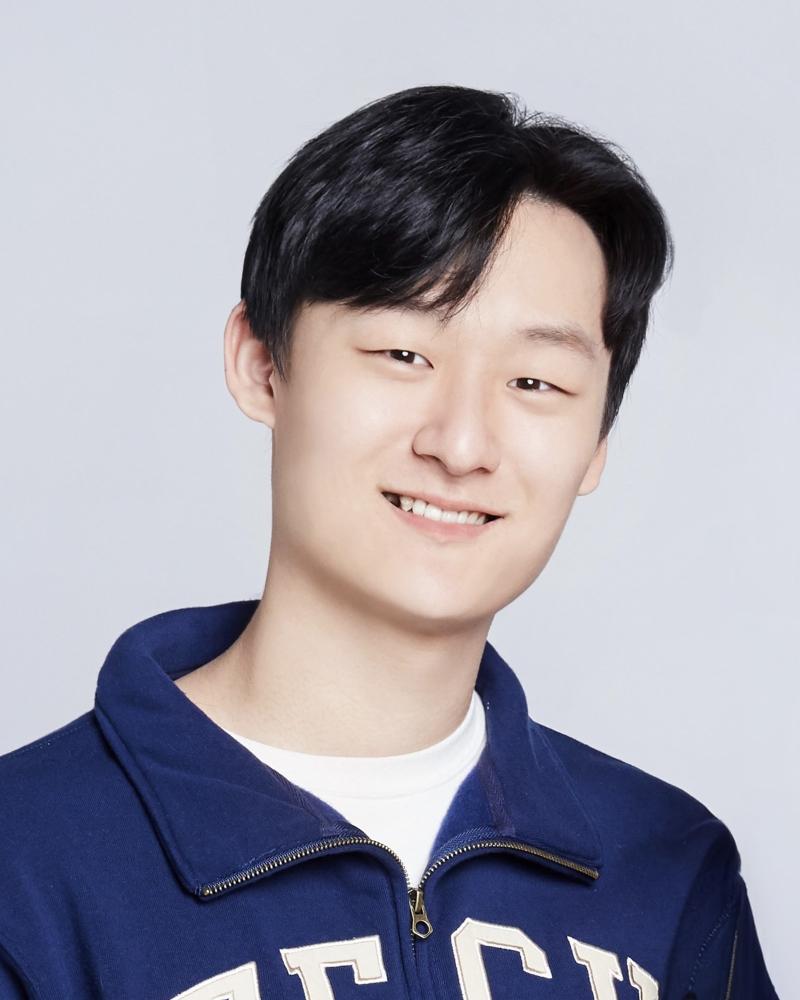}}]{Hyukbin Kwon} (S 2023) received the B.S. degree and is currently pursuing the Ph.D. degree at the G. W. Woodruff School of Mechanical Engineering at the Georgia Institute of Technology. His research interests include secure control systems and sensor fusion for robotic manipulation.
\end{IEEEbiography}

\end{document}